%% file: neurips_2026.tex
\documentclass{article}
\usepackage[main, final]{neurips_2026}
\usepackage[utf8]{inputenc} 
\usepackage[T1]{fontenc}    
\usepackage{hyperref}       
\usepackage{url}            
\usepackage{booktabs}       
\usepackage{amsfonts}       
\usepackage{nicefrac}       
\usepackage{microtype}      
\usepackage{xcolor}         
\usepackage[table]{xcolor}
\usepackage{algorithm}
\usepackage{algpseudocode}
\usepackage{graphicx}
\usepackage{multirow}
\usepackage{amsmath}
\usepackage{enumitem}
\usepackage{subcaption}
\usepackage{tikz}
\usepackage{hyperref}
\usepackage{wrapfig}
\usepackage[utf8]{inputenc} 
\usepackage{etoc} 

\hypersetup{
    colorlinks=true,
    linkcolor=blue,
    citecolor=blue,   
    urlcolor=blue      
}
\title{PGID: Progressive Guided Inversion and Denoising for Robust Watermark Detection}

\author{%
  Minh Quoc Duong \quad
  Chun Tong Lei \quad
  Chun Pong Lau\thanks{Corresponding author.} \\
  City University of Hong Kong \\
  \texttt{\{mduong2-c, ctlei2-c\}@my.cityu.edu.hk, cplau27@cityu.edu.hk}
}

\begin{document}

\maketitle

\input{sec/0_abstract}
\input{sec/1_intro}

\input{sec/2_related_works}

\input{sec/3_background}

\input{sec/4_method}

\input{sec/5_experiments}

\input{sec/6_conclusion}

\bibliographystyle{unsrt}
\bibliography{main}

\newpage
\input{sec/7_appendix}


\end{document}

%% file: sec/0_abstract.tex
\begin{abstract}
    With the proliferation of AI-generated images, digital watermarking has become an essential safeguard for protecting intellectual property and mitigating malicious exploitation. Recent works on semantic watermarking have enabled efficient copyright protection for diffusion models. However, the dependence of semantic watermarking on diffusion inversion for watermark detection creates a critical vulnerability. Imprint removal and forgery attacks exploit this weakness to produce deceptive results. Our analysis reveals that these attacks succeed by displacing watermarked latents into the \textit{unwatermarked region}, while guiding unwatermarked latents into the \textit{watermarked region}. Based on that, we propose \textbf{Progressive Guided Inversion and Denoising (PGID)}, the first plug-and-play, training-free noise extraction framework designed to defend against both attack strategies. PGID effectively defends by projecting perturbed latents back to the region where they originally belong. The projection is achieved by eliminating intermediate latent deflections and mitigating adversarial perturbations through progressive inversion-denoising cycles. Comprehensive evaluations across multiple schemes demonstrate that PGID successfully restores detection reliability by recovering removed watermarks and identifying forged instances.
    
\end{abstract}

%% file: sec/1_intro.tex
\section{Introduction}
Recent advances in diffusion models~\cite{rombach2022high, podell2023sdxl, esser2024scalingrectifiedflowtransformers} have empowered users to generate high-fidelity images at an unprecedented scale effortlessly, leading to a surge in AI-generated digital content. However, with this massive growth comes an increasing range of risks. The malicious use of generative models to spread disinformation and misguide communities presents a profound societal threat. Furthermore, without safeguards, service providers become vulnerable to intellectual property theft, unauthorized exploitation, and severe reputational damage.

\begin{figure}[t]
    \centering
    \subcaptionbox{\label{fig:teaser}}%
    {\includegraphics[height=4.8cm]{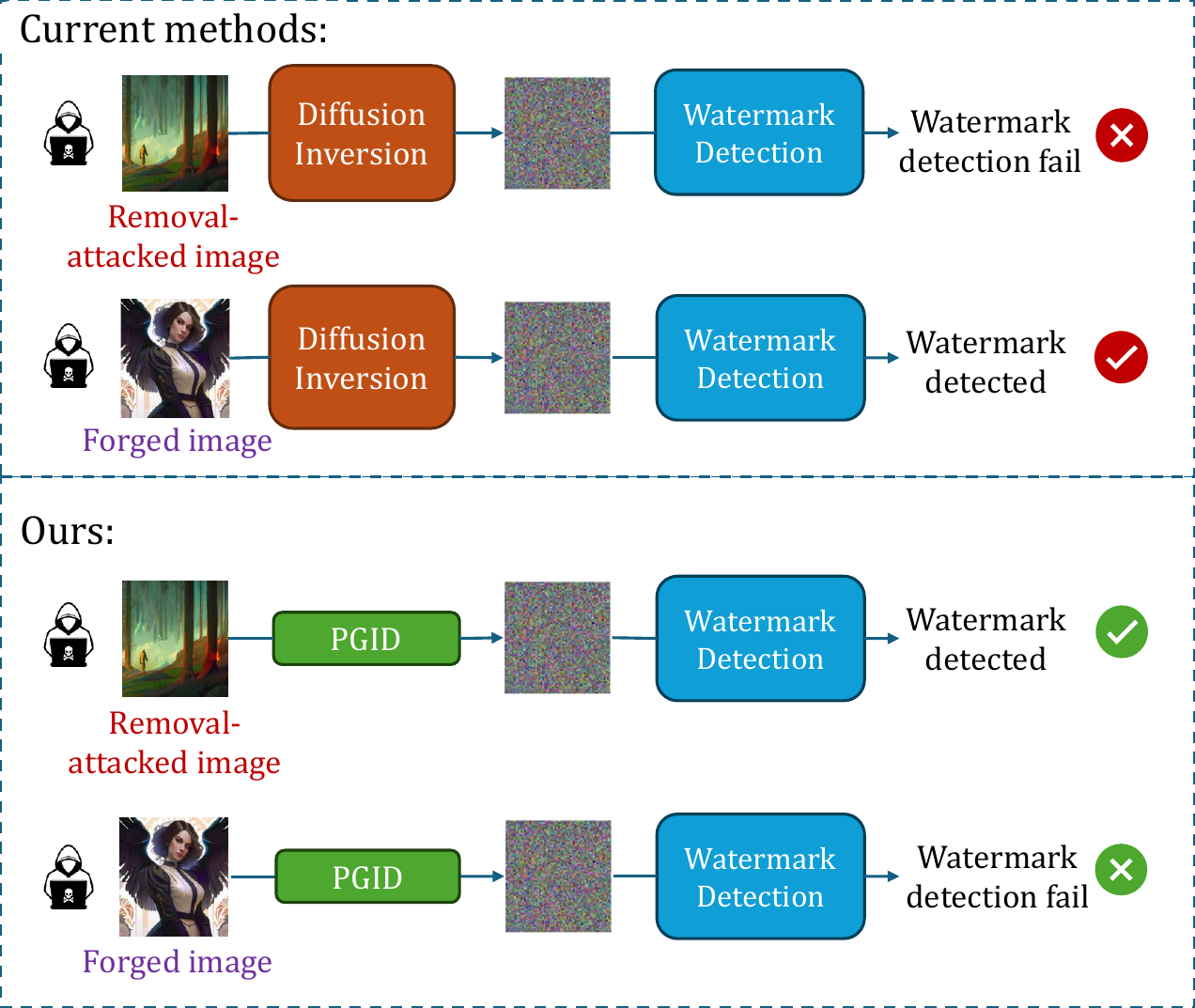}}
    \subcaptionbox{\label{fig:scenario}}%
    {\includegraphics[height=4.8cm]{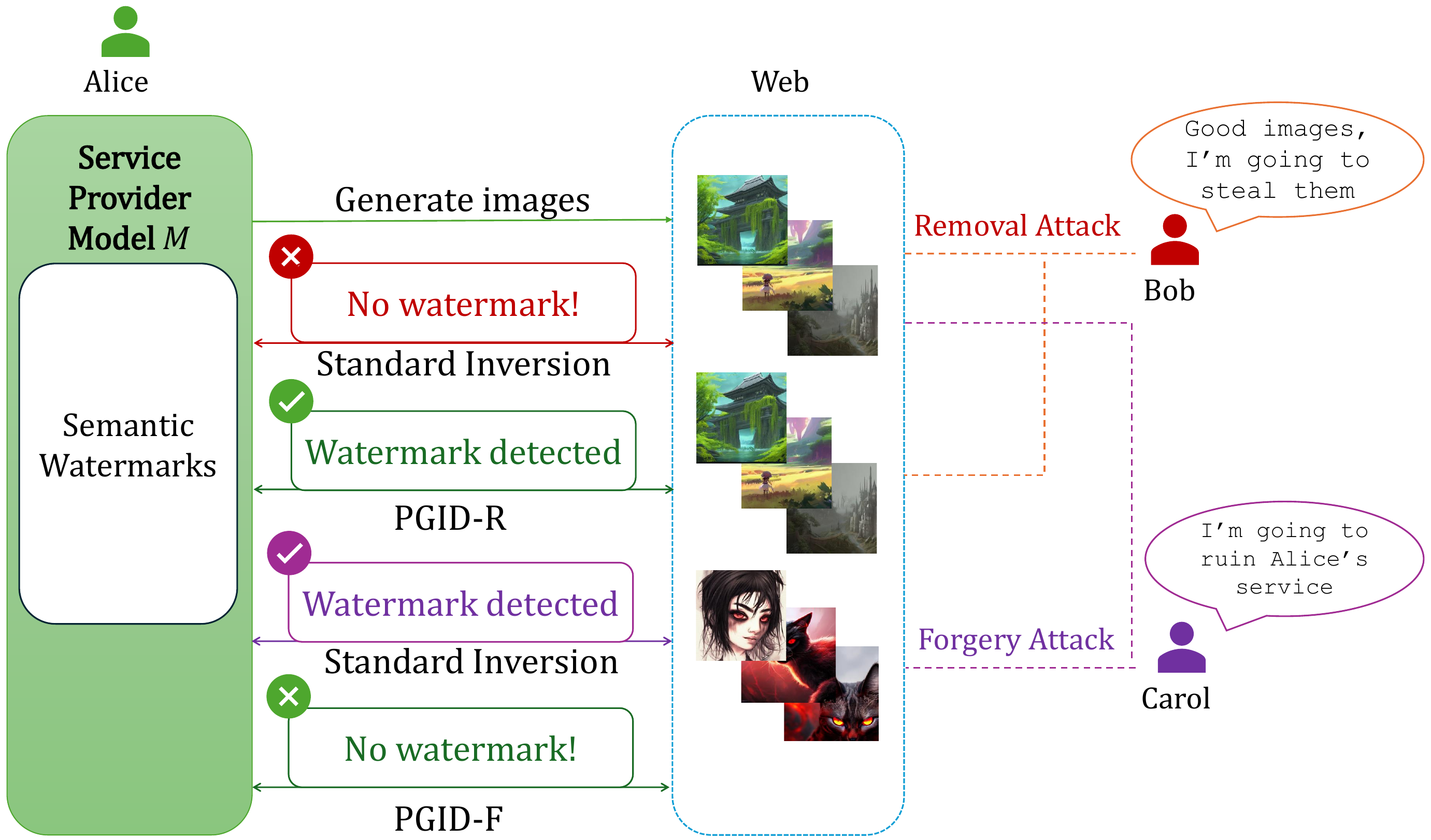}}
    
    \caption{Overview of the PGID framework and threat model. (a) Current semantic watermark methods are vulnerable to imprint forgery and removal attacks. PGID can recover watermark signals and reject forged images, providing effective defense. (b) The application scenario for PGID.}
    \label{fig:combined_teaser}
    \vspace{-0.7cm}
\end{figure}

Watermarking is an effective solution to mitigate the above risks by enabling AI-generated content detection, copyright authentication, and provenance tracing. In recent years, semantic watermarks~\cite{wen2023treering, yang2024gaussian, ci2024ringid, arabi2025seal, Muller_2025_CVPR} emerged as a state-of-the-art watermarking paradigm for diffusion models. These methods function by injecting watermark signals into the initial noise of the diffusion process and use it as input for image generation. Watermark extraction is achieved by tracing the generative process backward through diffusion inversion to retrieve an estimate of the initial noise. This recovered noise is subsequently decoded for watermark detection. 

Even though different semantic watermark schemes have different ways of injecting and decoding the embedded signals, they all share the same dependence on an accurate inversion function for initial noise retrieval, which is typically DDIM~\cite{songdenoising}. Exploiting this dependency, Müller et al. (2025)\cite{Muller_2025_CVPR} have demonstrated the vulnerability of this watermarking paradigm to black-box imprint forgery and removal attacks. Consequently, attackers can either falsely attribute illicit content to service providers through forgery or bypass accountability measures entirely by removing watermarks. This highlights a need for more robust extraction frameworks to complement current one-stage inversion approaches~\cite{osi2026onestep, Wang_2025_ROAR}.

We first observe that for any fixed watermark signal in an initial watermarked noise $z^*_T$, the latent space is partitioned into a \textit{watermarked region} and an \textit{unwatermarked region}. Under this formulation, removal attacks displace watermarked latents into the unwatermarked region, while forgery attacks guide cover images into the watermarked region. Based on that observation, we propose \textbf{Progressive Guided Inversion and Denoising (PGID)}, the first plug-and-play training-free noise extraction framework for defense against both imprint forgery and removal attacks. Our results show that, under the proposed unified framework, one can: (1) recover watermarks from images subjected to removal attack, and (2) differentiate between forged and authentic watermarked images. In our work, we refer to the usage of PGID against the removal attack and forgery attack as PGID-R and PGID-F, respectively. 

Fundamentally, our method projects the attacked latent back to the region it originally was in. This is achieved from the observation that imprint attacks induce cumulative distributional bias at every intermediate timestep. Our framework eliminates these deflections by traversing the diffusion trajectory through asymmetric cycles where \textbf{we denoise more than we invert} at each cycle, effectively mitigating the introduced perturbations while projecting the latent representation back toward its intended manifold. 

In summary, our contributions in this paper are:
\begin{itemize}[itemsep=2pt, topsep=2pt, parsep=0pt]
    \item We formalize the concepts of watermarked and unwatermarked regions for an initial noise watermark signal and empirically demonstrate that imprint attacks displace the latent representation out of its respective region to disrupt detection. 
    \item From there, we propose \textbf{Progressive Guided Inversion and Denoising (PGID)}, the first plug-and-play training-free noise extraction framework unifying both imprint forgery and removal defense for semantic watermarks.
    \item We validate the effectiveness of PGID with cutting-edge semantic watermarking schemes. Empirical results demonstrate that PGID is effective across different watermarking schemes and different attack settings. In the removal case, PGID can reliably recover the obscured watermark. To the best of our knowledge, it is the first framework capable of such a restoration.
\end{itemize}

%% file: sec/2_related_works.tex
\section{Related Works}

\subsection{Image Watermarking for Diffusion Models}

Image watermarking for diffusion models can be categorized into two primary methodologies: \textbf{post-processing} and \textbf{in-processing}. Post-processing methods apply watermarks directly to the final image using traditional transform-domain techniques~\cite{cox2008digital} or deep learning-based encoders~\cite{zhang2019robust,jia2021mbrs,ma2022towards}. However, these methods often compromise image fidelity, do not guarantee undetectability, and offer limited robustness against removal~\cite{yang2024steganalysisdigitalwatermarkingdefense, an2024benchmarking}. In-processing methods can be further divided into two subtypes: fine-tuning schemes and semantic watermarking schemes. Fine-tuning methods represent a more integrated approach where the watermark is embedded directly into the model’s parameters, typically by fine-tuning the Variational Autoencoder (VAE) decoder \cite{fernandez2023stable} or the noise predictor \cite{min2024watermark}. While this ensures all outputs are traceable and maintains high image quality, it incurs significant computational costs. Semantic watermarks remove the need for extensive model retraining by manipulating the initial input noise to a diffusion model and recovering the watermark by inverting the diffusion process. Techniques such as Tree Ring~\cite{wen2023treering} and RingID~\cite{ci2024ringid} utilize Fourier-space structuring to ensure geometric invariance, while methods like Gaussian Shading~\cite{yang2024gaussian} and PRC~\cite{gunn2025undetectable} bias the initial latent distribution to achieve efficient watermarking. In our work, we specifically focus on this semantic watermarking paradigm.

\subsection{Watermark Forgery and Removal}

As watermarking techniques advance, a parallel field of research has emerged focusing on the vulnerabilities of these systems to removal and forgery. The removal of post-processing image watermarks has been extensively investigated through various techniques, ranging from traditional photometric and degradation distortions ~\cite{hartung1999multimedia, Voloshynovskiy2001AttacksOD} to more sophisticated deep learning-based methods~\cite{an2024benchmarking, he2025transferattackimagewatermark}. Furthermore, recent advancements have introduced regeneration and reconstruction approaches~\cite{zhao2024invisible, liu2024ctrlregen} that leverage diffusion-based purification~\cite{nie2022diffusion, lei2025instant}. Likewise, forgery attacks on these traditional schemes are also well studied, including training a surrogate generative model or a preference model to estimate the watermark~\cite{dong2025wmcopierforginginvisibleimage, 2025transferableblackboxoneshotforging}. Most semantic watermark strategies have shown strong robustness against these adversaries~\cite{WINDwatermark2025, arabi2025seal, yang2025t2smark}. However, recently proposed imprint forgery and removal attacks~\cite{Muller_2025_CVPR} have highlighted a security gap in the paradigm's dependence on diffusion inversion. There have been works attempting to improve robustness against forgery attacks by binding the watermarks to image semantics~\cite{arabi2025seal, zhang2026sembind}, but these remain susceptible to removal. Our method fills the gap by tackling both removal and forgery attacks simultaneously within a unified framework.

%% file: sec/3_background.tex
\section{Preliminaries}
\subsection{Latent Diffusion Models (LDMs) and DDIM}
Diffusion models~\cite{ho2020denoising, song2019generative, song2021maximum, song2021scorebased, songdenoising} are generative models that are defined by a \textit{forward process} and a parameterized \textit{denoising process}. LDMs~\cite{rombach2022high} are diffusion models that operate in a compressed latent space rather than the pixel domain to achieve high-fidelity generation with computational efficiency. Given an image $x \sim q(x)$, an LDM first maps $x$ to a latent representation $z_0 = \mathcal{E}(x)$ via an encoder $\mathcal{E}$. Then the \textit{forward process} progressively degrades this latent $z_0$ by adding Gaussian noise to generate
a sequence of noisy samples $\{{z}_t\}^T_{t=1}$. At any given timestep $t \in [1, T]$, the noisy latent $z_t$ is defined as a linear combination of the initial latent and standard Gaussian noise $\epsilon \sim \mathcal{N}(0, I)$:

\begin{equation}
    z_t = \sqrt{\bar{\alpha}_t} z_0 + \sqrt{1-\bar{\alpha}_t} \epsilon
    \label{eq:forward}
\end{equation}

where $\bar{\alpha}_t = \prod_{i=1}^t (1 - \beta_i)$ is the cumulative product of a predefined variance schedule $\{\beta_t\}_{t=1}^T \in (0, 1)$.

The \textit{denoising process} aims to reverse this degradation to generate new data. It starts by sampling $z_T$ from the standard Gaussian distribution $p({z}_T) = \mathcal{N}({z}_T; 0,I)$. A parameterized neural network $\epsilon_\theta(z_t, t)$ is trained to estimate the added noise $\epsilon$ that was added at timestep $t$. A sampler then uses $\epsilon_\theta(z_t, t)$ to compute a less noisy latent $z_{t-1}$ from $z_{t}$. Numerous samplers have been proposed, and Denoising Diffusion Implicit Models (DDIM)~\cite{songdenoising} is among the most common methods because of its fast sampling speed and deterministic nature. With deterministic DDIM, a sampling step from $t$ to $t-1$ is defined as: 

\begin{equation}
    z_{t-1} = \sqrt{\bar{\alpha}_{t-1}} \left( \frac{z_t - \sqrt{1-\bar{\alpha}_t} \epsilon_\theta(z_t, t)}{\sqrt{\bar{\alpha}_t}} \right) + \sqrt{1-\bar{\alpha}_{t-1}} \epsilon_\theta(z_t, t)
    \label{eq:ddim}
\end{equation} 

An image is then generated by iteratively applying this deterministic update from $t=T$ down to $t=1$ to obtain the clean latent $z_0$, which is subsequently mapped back into the pixel space using a decoder $\mathcal{D}$, such that the final image is $x' = \mathcal{D}(z_0)$.

\subsection{Inverse DDIM}
Because DDIM is deterministic, one can follow the trajectory backwards from timestep $0$ up to $T$ to estimate the initial noise $z_T$. Given a generated image $x'$, we can first map it back to the latent domain $z_0 = \mathcal{E}(x')$. The Inverse DDIM process steps forward in time from $t$ to $t+1$ as follows:
\begin{equation}
    z_{t+1} = \sqrt{\bar{\alpha}_{t+1}} \left( \frac{z_t - \sqrt{1-\bar{\alpha}_t} \epsilon_\theta(z_t, t)}{\sqrt{\bar{\alpha}_t}} \right) + \sqrt{1-\bar{\alpha}_{t+1}} \epsilon_\theta(z_t, t)
    \label{eq:inv_ddim}
\end{equation}

For our work, we denote a single step of DDIM sampling (from timestep $t \text{ to } t-1$) and inverse DDIM (from timestep $t \text{ to } t+1$) as ${Denoise}(z, t)$ and ${Inverse}(z, t)$, respectively. Let $\mathcal{T}_{0 \to T}(z_0)$ represent the inversion process that yields the latent trajectory $\{z_t\}_{t=0}^T$, and $I_{0 \to T}(z_0)$ denote the mapping to the final estimated noise $z_T$.

%% file: sec/4_method.tex
\begin{figure}[t]
    \centering
    
    \subcaptionbox{\label{fig:gs_pca}}%
    {\includegraphics[width=0.35\linewidth]{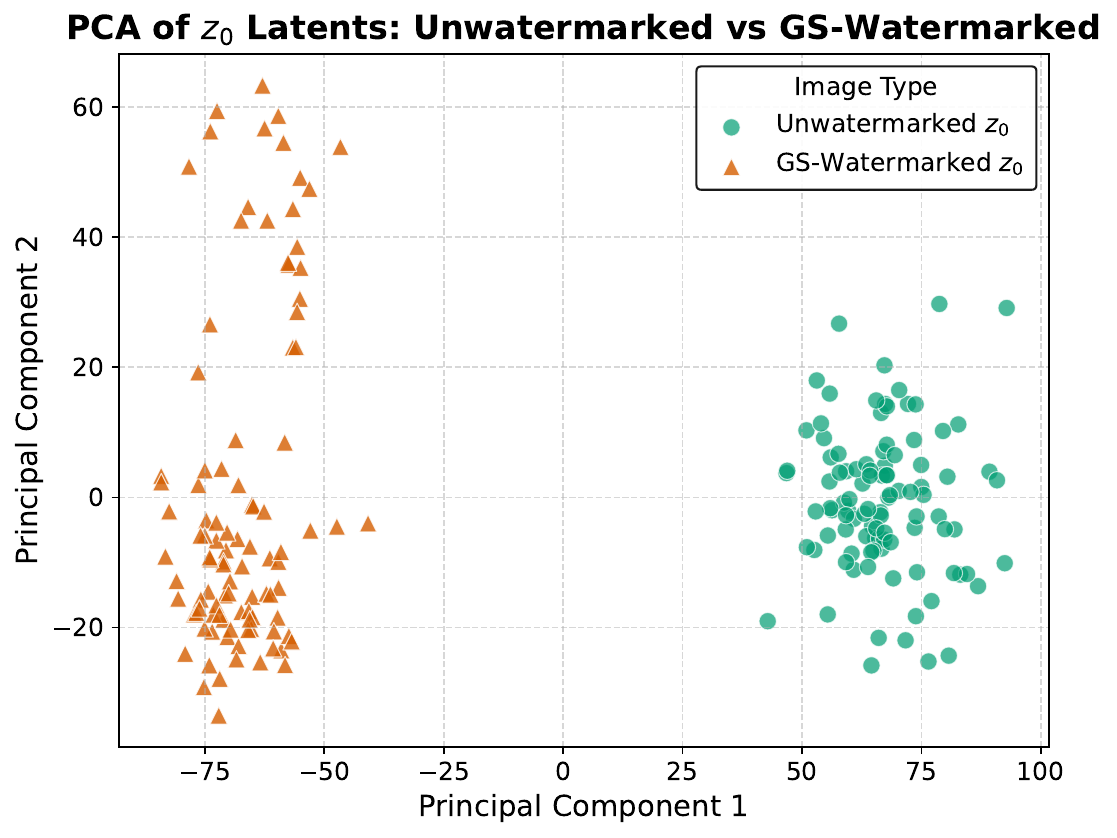}}%
    \hspace{0.5cm}
    \subcaptionbox{\label{fig:tr_pca}}%
    {\includegraphics[width=0.35\linewidth]{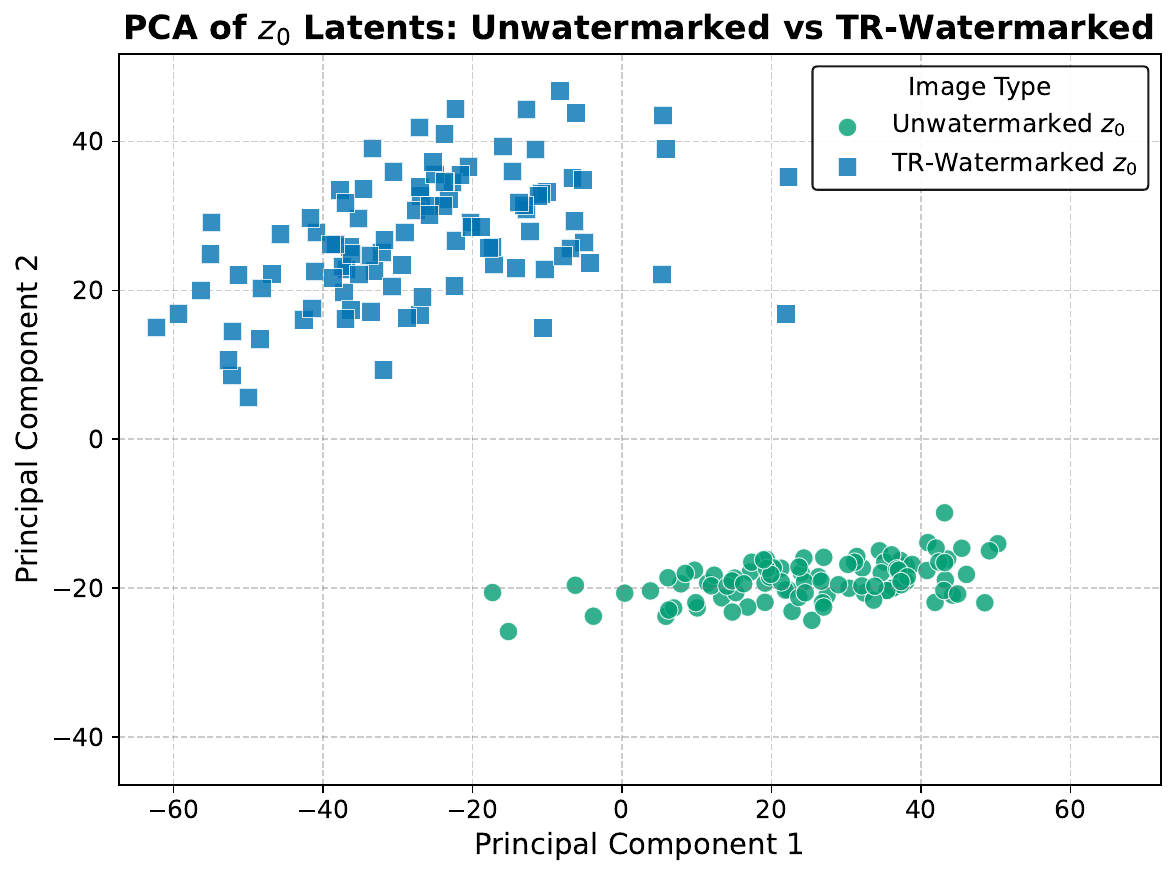}}
    \caption{Visualization of latent manifold separation. PCA projections of the latent space $\mathcal{Z}_0$ for (a) Gaussian Shading and (b) Tree Ring. The visualization demonstrates a clear structural differentiation between the watermarked region and the unwatermarked region for a fixed initial noise signal.} 
    \label{fig:combined_pca_ab}
    \vspace{-0.55cm}
\end{figure}
\section{Method}
In this section, we describe how service providers can utilize PGID to protect themselves from imprint forgery and removal attacks. Subsequently, we formalize the latent-space partition into watermarked and unwatermarked regions given a watermark signal, and describe the corresponding behavior of removal and forgery attacks in this space. Built upon that, we then present PGID in detail.

\subsection{Application Scenarios}

\textbf{Scenarios.} We consider a scenario in which Alice provides an API for users to access her generative model $M$. The model $M$ is kept hidden and closed-source from users. Alice employs a semantic watermarking method for copyright protection and traceability. However, Alice faces threats from two types of malicious users: Bob, who seeks to steal Alice's generated images by performing an imprint removal attack to claim them as his own, and Carol, who seeks to damage Alice's reputation by using an imprint forgery attack to embed Alice's watermark into unauthorized or illicit images. If Alice relies solely on a standard extraction pipeline, both attacks succeed: her watermark is not detected in Bob's stolen images, whereas it is falsely detected in Carol's forged images. Consequently, Alice is left vulnerable to copyright loss and potential legal liability. 

\textbf{Usage. }Given an arbitrary image, Alice does not know whether it has undergone a removal attack, a forgery attack, or no attack at all. To counteract both threats, Alice integrates PGID, which requires only her model $M$, into her system. For any suspect image, Alice first runs the standard watermark extraction, using its outcome to dictate her next step. \textit{(1) Verifying Rejections (Defending against Bob):} If the standard detection yields a \textit{negative} result, but Alice suspects the image was stolen from her service, she applies PGID-R. If PGID-R successfully recovers the hidden watermark, she confirms a removal attack occurred and reclaims her copyright. \textit{(2) Verifying Authenticity (Defending against Carol):} If the standard detection yields a \textit{positive} result on a suspicious or illicit image, she applies PGID-F. If the watermark is no longer detected, she confirms that it was a forged instance. The application scenario is presented in Figure~\ref{fig:scenario}.

\subsection{Watermarked and Unwatermarked Latent Regions}
\textbf{Formalization. }Jain et al. (2025)\cite{jain2025forging} first proposed the existence of a \textit{watermarked region} for forgery exploitation. In our work, we further formalize and empirically show that such a region and its complement, the \textit{unwatermarked region}, exist with respect to the specific watermark signal embedded in an \textbf{initial watermarked noise $z_T^*$}. Given a watermarking method and an initial watermarked noise $z_T^*$, let $\mathcal{D}_{z_T^*}: \mathcal{Z}_T \to \{0, 1\}$ be a binary detection function that evaluates whether a given noise vector contains the specific watermark signal in $z_T^*$. We formalize the watermarked region $\mathcal{R}_{w}(z_T^*)$ as the set of all latents $z_0$ that invert to a noise vector triggering a successful detection:
$$\mathcal{R}_{w}(z_T^*) = \{ z_0 \in \mathcal{Z}_0 \mid \mathcal{D}_{z_T^*}(I_{0\to T} (z_{0})) = 1 \}$$
In other words, there exist many $z_0$ mapping to a noise $z_T$ through the inversion process that contains the specific watermark signal from $z^*_T$, which collectively constitute the watermarked region $\mathcal{R}_{w}(z_T^*)$.
Conversely, we define the \textit{unwatermarked region} $\mathcal{R}_{u}(z_T^*)$ as the complement of the watermarked region in the latent space. It comprises all latents $z_0$ whose inverted noise fails to yield a detectable signal for $z_T^*$:
$$\mathcal{R}_{u}(z_T^*) = \{ z_0 \in \mathcal{Z}_0 \mid \mathcal{D}_{z_T^*}(I_{0\to T} (z_{0})) = 0 \}$$
Equivalently, $\mathcal{R}_{u}(z_T^*) = \mathcal{Z}_0 \setminus \mathcal{R}_{w}(z_T^*)$. 
    We empirically show that these regions exist by performing a toy experiment on Tree Ring and Gaussian Shading, where we inject only a single watermark signal across watermarked images. For Tree Ring, we inject only one key pattern across all images (this is the default mechanism of Tree Ring). For Gaussian Shading, we generate all images using a single key and message. Consequently, if we fix the initial noise of any single watermarked image as $z_T^*$, the latents $z_0$ of the remaining watermarked images belong to the corresponding watermarked region $\mathcal{R}_w(z_T^*)$. Under this setup, we perform PCA on $100$ watermarked images and $100$ unwatermarked images from the two schemes. From Figure~\ref{fig:combined_pca_ab}, the watermarked and unwatermarked regions are clearly separated when conditioned on a specific watermark signal, confirming our hypothesis. 
\input{algo/PGID}
\begin{figure}[t]
    \centering
    \subcaptionbox{\label{fig:tr_pca_all}}%
    {\includegraphics[width=0.35\linewidth]{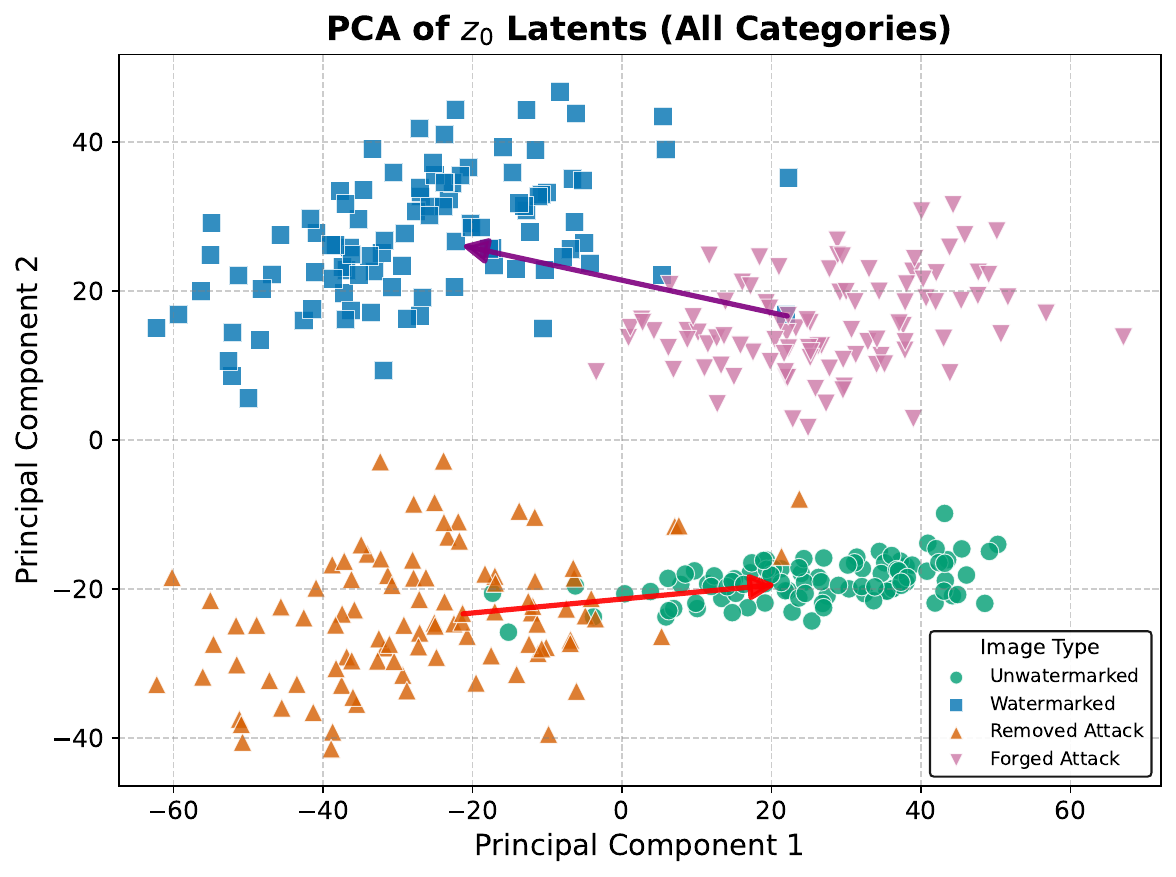}}
    \hspace{0.5cm}
    \subcaptionbox{\label{fig:tr_pca_after}}%
    {\includegraphics[width=0.35\linewidth]{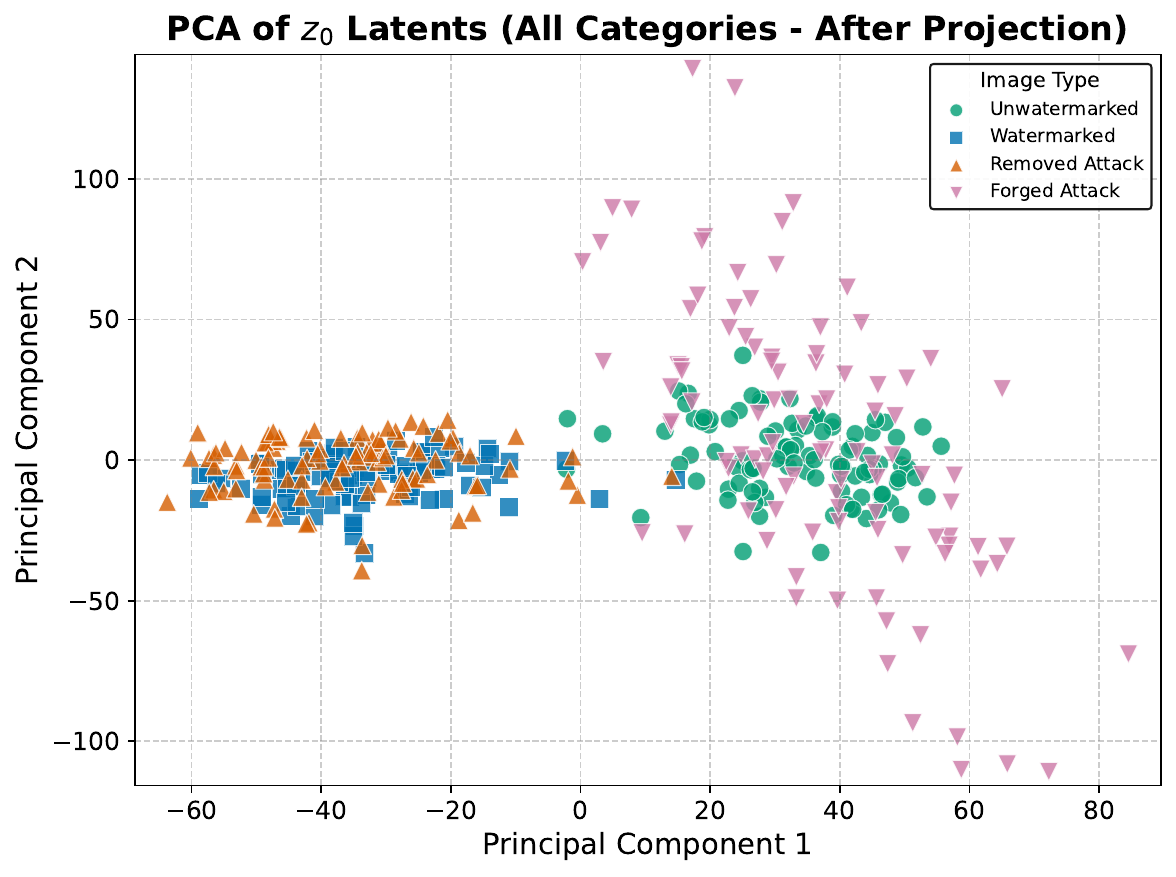}}
    
    \caption{(a) PCA visualization of latent-space clean and attacked regions for Tree Ring. Forgery guides cover latents toward the watermarked region, while removal attacks displace watermarked latents into the unwatermarked region. (b) PCA visualization of latent-space clean and attacked regions for Tree Ring after projection by PGID. It can be observed that attacked latents are projected back to their original regions.}
    \label{fig:combined_pca_analysis}
    \vspace{-0.55cm}
\end{figure}

\textbf{Threat Model: Removal and Forgery. }Let $x^{(w)}$ and $x^{(c)}$ be a watermarked image generated from initial watermarked noise $z^*_T$ and an unwatermarked cover image, respectively. Let $z^{(w)}_0 = \mathcal E(x^{(w)})$ and $z^{(c)}_0 = \mathcal E(x^{(c)})$ where $\mathcal E$ is the encoder from the service provider's model $M$.
Under this formulation, an imprint removal attack can be viewed as finding a perturbation $\delta$ that shifts an originally watermarked latent into the unwatermarked region ($z^{atk}_{0}= z^{(w)}_0 + \delta \in \mathcal{R}_{u}(z_T^*)$). Conversely, an imprint forgery attack constructs a perturbation to force a cover latent into the watermarked region ($z^{atk}_{0}= z^{(c)}_0 + \delta \in \mathcal{R}_{w}(z_T^*)$). We show this by visualizing PCA of watermarked, unwatermarked, removal-attacked, and forged latents from Tree Ring. As illustrated in Figure~\ref{fig:tr_pca_all}, forged latents converge toward the watermarked region, while removal-attacked latents are displaced into the unwatermarked distribution.

\subsection{Progressive Guided Inversion and Denoising (PGID)}
From the above results, a valid defense strategy against the removal attack is to project $z_0^{atk}$ back to an instance in the watermarked region, $z_0 \in \mathcal{R}_w(z_T^*)$. Likewise, against forgery, we can project $z_0^{atk}$ back to the unwatermarked region $z_0 \in \mathcal{R}_u(z_T^*)$. This intuition is achieved by our PGID method.

\textbf{Overview. }PGID can be divided into three stages as described in Algorithm~\ref{algo:PGID}. The first stage involves running the standard DDIM inversion to obtain attacked intermediate latents $\{z^{atk}_t\}_{t=0}^T$. The second stage is where we perform progressive latent refinement and obtain the projected $z_0$ by utilizing the precomputed intermediate latents from the first stage. The final stage is running standard inversion with the projected $z_0$ to obtain the final noise $z_T$. 

\begin{wrapfigure}{r}{0.4\textwidth}
    \centering
    \includegraphics[width=\linewidth]{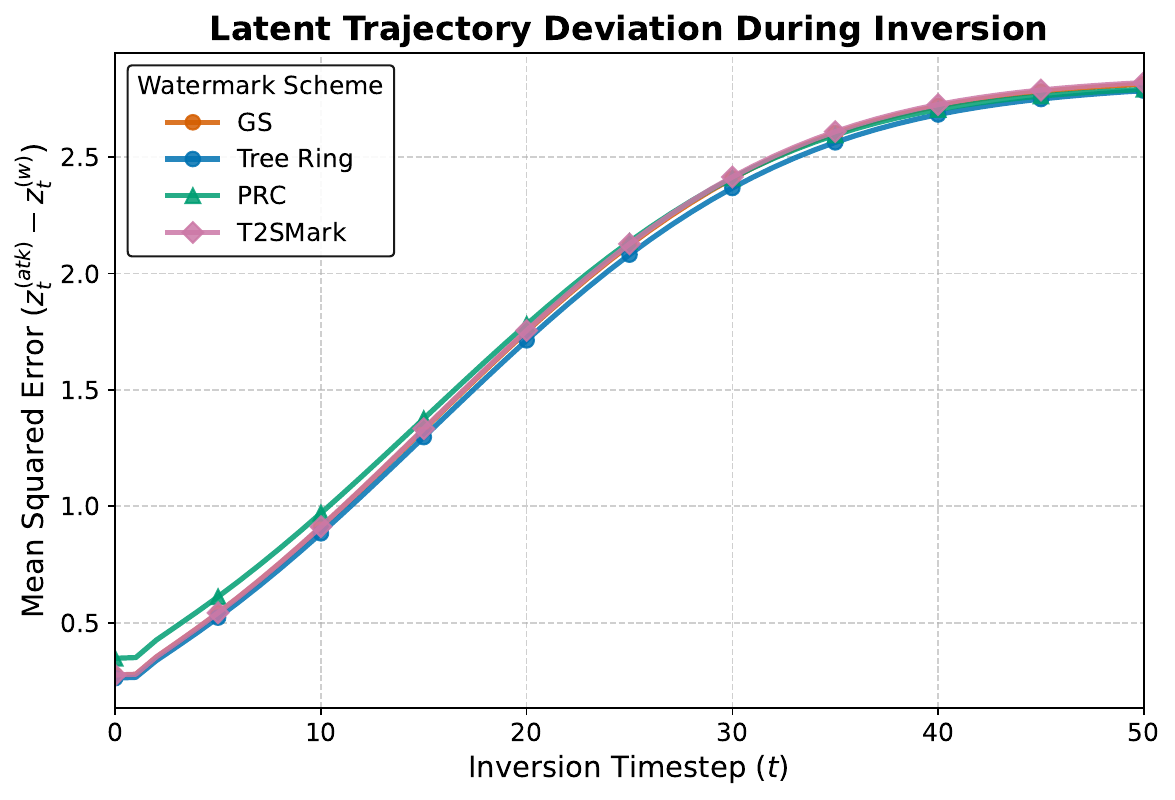}
    \caption{Visualization of the trajectory deviation MSE caused by the removal attack from $t=0$ to $t=50$. The inversion discrepancy accumulates across intermediate timesteps.}
    \label{fig:mse_trajectory}
\end{wrapfigure}

\textbf{Progressive Latent Refinement. }We focus on the second stage of our framework, where we progressively refine our latent projection. As described in Figure~\ref{fig:mse_trajectory}, we found that the final inversion error caused by the attacks is cascaded through intermediate timesteps. To fix the errors caused by this, we guide the inversion trajectory of the projected latent away from the precomputed $\{z^{atk}_t\}_{t=0}^T$ during inversion. Then, by denoising back, using the model's clean manifold projection capability, the latent has its inversion trajectory refined. To further remove the effects of the perturbation, we skip some inversion steps while keeping the same denoising steps every cycle. The intuition behind this process is to treat the introduced adversarial perturbation as additional noise on the original clean latent. Therefore, by inverting fewer steps than denoising, we can project the perturbed latent to the clean manifold more effectively. This stage uses three hyperparameters: the stopping timestep $k < T$, the number of skip inversions $s$, and guidance strength $\gamma$. The stopping timestep $k$ confines our progressive cycle to operate on the first $k$ timesteps, $s$ determines how many inversion steps we skip in each cycle, while $\gamma$ dictates the intensity of trajectory guidance. With different sets of these hyperparameters, PGID can be used to effectively defend against both removal and forgery. Figure~\ref{fig:tr_pca_after} shows that PGID successfully projects attacked latents back to their original regions: removal-attacked back to watermarked and forged back to unwatermarked.

%% file: algo/PGID.tex
\begin{algorithm}[t]
\caption{PGID}
\label{algo:PGID}
\begin{algorithmic}[1]
\Require Suspect image $x_{atk}$, stopping timestep $k$, number of skip inversion steps $s$, guidance strength $\gamma$, and model $M$ used for watermarked image generation with encoder $\mathcal{E}$ and decoder $\mathcal{D}$.

\Statex \textbf{Stage I: Trajectory Pre-computation}
\State $z^{atk}_0 \gets \mathcal{E}(x^{atk})$
\State Let $\{z^{atk}_t\}_{t=0}^T \gets \mathcal{T}_{0 \to T}(z_0^{atk})$ \Comment{Pre-compute the standard inversion trajectory}

\Statex \textbf{Stage II: Progressive Latent Refinement}
\State $\hat{z} \gets z_0^{atk}$ \Comment{Initialization}
\For{$i = 1$ to $k$}
    \For{$t = 1$ to $i - s$} \Comment{Invert $i-s$ steps}
        \State $\hat{z} \gets Inverse(\hat{z}, t-1)$ 
        \State $\hat{z} \gets \hat{z} + \gamma(\hat{z}-z^{atk}_t)$
    \EndFor
    
    \For{$t = i$ down to $1$} \Comment{Denoise $i$ steps}
        \State $\hat{z} \gets Denoise(\hat{z}, t)$ 
    \EndFor
\EndFor
\State $z_0 \gets \hat{z}$ \Comment{Finish refinement}
\Statex \textbf{Stage III: Final Inversion}
\State $z_T \gets I_{0\to T}(z_0)$ \Comment{Final full inversion}
\State \Return $z_T$
\end{algorithmic}
\end{algorithm}

%% file: sec/5_experiments.tex
\section{Experiments}
\subsection{Experimental Setting}
In our main experiments, we employ \textit{Stable Diffusion v2.1} (SD 2.1)\footnote{https://huggingface.co/Manojb/stable-diffusion-2-1-base} as our service provider model. All images are generated at a resolution of $512\times512$ using classifier-free guidance with a scale of $7.5$ and $50$ inference steps. All images are generated with prompts from the Stable Diffusion Prompts\footnote{https://huggingface.co/datasets/Gustavosta/Stable-Diffusion-Prompts} (SDP) dataset. All of our experiments are run on an NVIDIA L40 GPU.

\input{tables/main_results_removal}
\input{tables/main_results_forgery}

\textbf{Watermarking schemes. }We evaluate PGID's effectiveness with four semantic watermarking schemes: Tree Ring~\cite{wen2023treering}, Gaussian Shading~\cite{yang2024gaussian}, PRC~\cite{gunn2025undetectable}, and T2SMark~\cite{yang2025t2smark}. For each scheme, we compare the performance of its default inversion process against its PGID-augmented variant (denoted "+P"). All schemes that allow embedded bit messages (Gaussian Shading, PRC, and T2SMark) are encoded with $256$ bits. Unless specified otherwise, we follow the default settings of the watermarking schemes for embedding and detection. Detailed settings for each method are in Appendix \ref{appendix:semantic_watermark}.

\textbf{Hyperparameter settings. } For PGID-R (against removal), we set the stopping timestep $k=10$, number of skip inversion steps $s=1$, and guidance strength $\gamma=0.045$ for all watermarking schemes. For PGID-F (against forgery), we employ $k=15, s=3, \gamma = 0.001$ for Gaussian Shading, Tree Ring, and T2SMark. The setting for PRC is the same, with the only difference being $k=7$. We set $T$ in stages I and III to the default $50$. We provide an analysis of different hyperparameter settings at Appendix~\ref{appendix:hyperparam_settings}.

\textbf{Imprint attacks. }Following Müller et al. (2025)~\cite{Muller_2025_CVPR}, we consider the imprint attacks with $50$, $100$ and $150$ optimization steps. In our main experiments, we use two models as the proxy attacker model: \textit{Stable Diffusion v2.1} (SD 2.1) and \textit{Stable Diffusion v1.5} (SD 1.5), representing the cases where the attacker's model matches and does not match the service provider model, respectively. By default, the DDIM scheduler is used as the attacker's sampler with $50$ inversion steps. For the imprint forgery attack, we report results on two different datasets as cover images: ImageNet~\cite{deng2009imagenet} and MS-COCO~\cite{lin2014microsoft}. In our main results, for each watermarking method, our evaluation is based on $100$ watermarked images, $100$ generated unwatermarked images, $100$ removal-attacked images, and $100$ forged images for each proxy model and cover image dataset.

\textbf{Evaluation. }We benchmark detection effectiveness using AUC and TPR at fixed FPR thresholds (denoted as Det.). To maintain consistency with the original literature, we set the FPR of $10^{-6}$ for Gaussian Shading, PRC, and T2SMark, while Tree Ring is evaluated at the FPR of $1\%$. In addition, we also report the bit accuracy, defined as the ratio of correctly reconstructed bits to the total payload length, for watermarking schemes that support multi-bit message embedding. Notably, since PRC utilizes image-dependent dynamic thresholds rather than a global one, its AUC calculation is omitted. In the removal case, the detection metrics are calculated between generated unwatermarked images and removal-attacked images, meaning higher AUC and TPR indicate higher robustness. Meanwhile, in the forgery case, the detection metrics are calculated between generated unwatermarked images and forged images. Higher AUC and TPR indicate more successful forgery attacks where the forged watermarks are highly distinguishable from the unwatermarked images. We provide more details of the thresholds and evaluations in Appendix~\ref{appendix:semantic_watermark}.

\subsection{Main Results}

\textbf{Against removal attack. }Table~\ref{tab:removal_attack} reports the watermark detection results in the removal case. Baseline schemes using standard inversion are highly vulnerable to the attack, leading to nearly none being able to detect the watermark after removal in both proxy model settings. T2SMark is the only watermarking scheme that remains fairly robust but loses all payload information ($\sim0.5$ bit accuracy). PGID-R effectively restores these obscured watermarks, achieving nearly 100\% detection and AUC across all schemes and attacker models. Furthermore, PGID-R also increases the bit accuracy to nearly perfect bit message retrieval, with the most substantial increase in Gaussian Shading: from roughly $0.00$ to nearly $1.00$. The AUC for all methods is also near $100\%$, showing that with PGID-R, originally watermarked images after removal attack can easily be distinguished from unwatermarked images.

\input{tables/main_results_forgery_imagenet}
\input{tables/watermarked}

\textbf{Against forgery attack. }Table~\ref{tab:forgery_attack} and Table~\ref{tab:forgery_attack_imagenet} show the results against forgery attacks. All baseline watermarking methods are extremely susceptible to the forgery attack, with nearly 100\% of forged images misidentified as watermarked. PGID-F suppresses the forged signal, reducing detection rates to near zero and bit accuracy to random-guess levels ($\sim0.5$). Generally, the AUC for all schemes also drops when using PGID-F. This shows that with PGID-F, the test statistics of forged images are re-aligned to be closer to those of unwatermarked images. In the case of Gaussian Shading, while the forged and unwatermarked distributions remain mathematically separable (reflected in a high AUC), most forged images are no longer detected as watermarked. This shows that PGID-F successfully degrades the forged imprint below the detection threshold, effectively securing the system.

\subsection{Authentic Watermark Robustness and Separability}
We further analyze our method's ability to separate authentic watermarks from forged ones by running watermarked images through PGID-F, as shown in Table~\ref{tab:watermarked_robustness}. We report the same metrics as the main experiment, with the only difference being the AUC. The AUC in Table~\ref{tab:watermarked_robustness} is calculated between watermarked images and forged images from MS-COCO and ImageNet. As shown, PGID-F consistently preserves the authentic watermarks, maintaining detection rates above $0.93$ and high bit accuracies across all schemes. Furthermore, the defense achieves near-perfect AUC scores against both SD 2.1 and SD 1.5 attacker models across all optimization steps. This demonstrates that our method can reliably isolate genuine watermarks from forged ones.
\input{tables/PixArt_both}
\subsection{Generalizability on Other Architectures}
\input{tables/PixArt_robustness}
To test the application of our method to other architectures, we utilize \textit{PixArt-$\alpha$}\footnote{https://huggingface.co/PixArt-alpha/PixArt-XL-2-512x512}~\cite{chen2023pixartalpha}, a model with a diffusion transformer backbone as our service provider model, and SD 2.1 as the attacker model. We report the same metrics as in the main experiments. The metrics are evaluated based on $30$ generated unwatermarked images, $30$ watermarked, and $30$ attacked images from each strategy. Cover images are taken from the MS-COCO dataset. We use a different hyperparameter setting from our main experiments. Details about our settings are provided in Appendix~\ref{appendix:hypersettings_in_paper}.

Results are presented in Table~\ref{tab:pixart_removal_attack}, Table~\ref{tab:pixart_forgery_attack} and Table~\ref{tab:pixart_watermarked_robustness}. Overall, PGID retains similar performance when applied to a DiT-based model. Against removal, PGID-R continues to restore the watermark signals with high accuracy, exhibiting only a modest performance degradation compared to the main results. In the forgery case, although there is a decrease in protection against attacks with higher optimization steps, PGID-F retains its capability to differentiate between forged and authentic watermarked instances, reflected by the high AUC depicted in Table~\ref{tab:pixart_watermarked_robustness}. These results highlight the plug-and-play and training-free features of our method.

%% file: tables/main_results_removal.tex
\begin{table}[t]
\centering
\caption{Results against imprint removal attack for four semantic watermarking schemes and their PGID-augmented variants. Results are reported for the attack at 50/100/150 optimization steps. Higher AUC and Det. indicate higher robustness to the attack. The best result for each scheme is in \textbf{bold}.}
\vspace{1ex}
\label{tab:removal_attack}
\setlength{\tabcolsep}{1mm}
\resizebox{\textwidth}{!}{
\begin{tabular}{@{}l ccc ccc@{}}
\toprule
\multirow{2}{*}{Method} & \multicolumn{3}{c}{Attacker Model: SD 2.1} & \multicolumn{3}{c}{Attacker Model: SD 1.5} \\
\cmidrule(lr){2-4} \cmidrule(l){5-7}
& Det.$\uparrow$ & Bit Acc.$\uparrow$ & AUC$\uparrow$ & Det.$\uparrow$ & Bit Acc.$\uparrow$ & AUC$\uparrow$ \\
\midrule
TR        & 0.24/0.13/0.11 & --- & 0.4085/0.1781/0.1363 & 0.25/0.13/0.10 & --- & 0.4235/0.1942/0.1393 \\
\rowcolor{gray!15}
TR+P      & \textbf{0.99}/\textbf{1.00}/\textbf{1.00} & --- & \textbf{0.9998}/\textbf{0.9988}/\textbf{0.9998} & \textbf{1.00}/\textbf{1.00}/\textbf{1.00} & --- & \textbf{1.0000}/\textbf{1.0000}/\textbf{0.9999}\\
GS        & 0.00/0.00/0.00 & 0.2582/0.0342/0.0077 & 0.0747/0.0000/0.0000 & 0.01/0.00/0.00 & 0.2891/0.0454/0.0114 & 0.1022/0.0000/0.0000\\
\rowcolor{gray!15}
GS+P      & \textbf{1.00}/\textbf{1.00}/\textbf{1.00} & \textbf{0.9975}/\textbf{0.9987}/\textbf{0.9981} & \textbf{1.0000}/\textbf{1.0000}/\textbf{1.0000} & \textbf{1.00}/\textbf{1.00}/\textbf{1.00} & \textbf{0.9964}/\textbf{0.9981}/\textbf{0.9968} & \textbf{1.0000}/\textbf{1.0000}/\textbf{1.0000} \\
PRC       & 0.00/0.00/0.00 & 0.5012/0.5038/0.5020 & --- & 0.00/0.00/0.00 & 0.5011/0.4980/0.5039 & --- \\
\rowcolor{gray!15}
PRC+P     & \textbf{0.97}/\textbf{1.00}/\textbf{1.00} & \textbf{0.9590}/\textbf{1.0000}/\textbf{0.9847} & --- & \textbf{0.97}/\textbf{1.00}/\textbf{0.98} & \textbf{0.8907}/\textbf{0.9841}/\textbf{0.9710} & ---\\
T2SMark   & 0.50/0.94/\textbf{1.00} & 0.5093/0.4977/0.4965 &  0.8695/0.9991/\textbf{1.0000} & 0.44/0.94/\textbf{1.00} & 0.5177/0.4964/0.4946 & 0.8806/0.9976/\textbf{1.0000}\\
\rowcolor{gray!15}
T2SMark+P & \textbf{1.00}/\textbf{1.00}/\textbf{1.00} & \textbf{0.9999}/\textbf{0.9999}/\textbf{1.0000} & \textbf{1.0000}/\textbf{1.0000}/\textbf{1.0000} & \textbf{1.00}/\textbf{1.00}/\textbf{1.00} & \textbf{0.9998}/\textbf{0.9999}/\textbf{1.0000} & \textbf{1.0000}/\textbf{1.0000}/\textbf{1.0000}\\
\bottomrule
\end{tabular}}
\end{table}

%% file: tables/main_results_forgery.tex
\begin{table}[t]
\centering
\caption{Results against imprint forgery attack on MS-COCO dataset for four semantic watermarking schemes and their PGID-augmented variants. Results are reported for the attack at 50/100/150 optimization steps. Lower AUC and Det. indicate higher robustness to the attack. The best result for each scheme is in \textbf{bold}.}
\vspace{1ex}
\label{tab:forgery_attack}
\setlength{\tabcolsep}{1mm}
\resizebox{\textwidth}{!}{
\begin{tabular}{@{}l ccc ccc@{}}
\toprule
\multirow{2}{*}{Method} & \multicolumn{3}{c}{Attacker Model: SD 2.1} & \multicolumn{3}{c}{Attacker Model: SD 1.5} \\
\cmidrule(lr){2-4} \cmidrule(l){5-7}
& Det.$\downarrow$ & Bit Acc.$\downarrow$ & AUC$\downarrow$ & Det.$\downarrow$ & Bit Acc.$\downarrow$ & AUC$\downarrow$ \\
\midrule
TR        & 1.00/1.00/1.00 & --- & 1.0000/1.0000/1.0000 & 1.00/1.00/1.00 & --- & 1.0000/1.0000/1.0000\\
\rowcolor{gray!15}
TR+P      & \textbf{0.02}/\textbf{0.04}/\textbf{0.07} & --- & \textbf{0.5472}/\textbf{0.6048}/\textbf{0.6728} & \textbf{0.04}/\textbf{0.08}/\textbf{0.09} & --- & \textbf{0.5515}/\textbf{0.6139}/\textbf{0.6625}\\
GS        & 1.00/1.00/1.00 & 0.9987/0.9994/0.9995 & 1.0000/1.0000/1.0000 & 1.00/1.00/1.00 & 0.9983/0.9991/0.9995 & 1.0000/1.0000/1.0000\\
\rowcolor{gray!15}
GS+P      & \textbf{0.00}/\textbf{0.00}/\textbf{0.00} & \textbf{0.5358}/\textbf{0.5764}/\textbf{0.6099} & \textbf{0.7773}/\textbf{0.9244}/\textbf{0.9582} & \textbf{0.00}/\textbf{0.00}/\textbf{0.06} & \textbf{0.5410}/\textbf{0.5849}/\textbf{0.6139} & \textbf{0.7969}/\textbf{0.9108}/\textbf{0.9577} \\
PRC       & 1.00/1.00/1.00 & 1.0000/1.0000/1.0000 & --- & 0.94/1.00/1.00 & 0.8788/0.9895/1.0000 & ---\\
\rowcolor{gray!15}
PRC+P     & \textbf{0.00}/\textbf{0.00}/\textbf{0.01} & \textbf{0.4952}/\textbf{0.5000}/\textbf{0.5021} & --- & \textbf{0.00}/\textbf{0.00}/\textbf{0.03} & \textbf{0.4980}/\textbf{0.5027}/\textbf{0.5000} & --- \\
T2SMark   & 1.00/1.00/1.00 & 0.9999/1.0000/1.0000 & 1.0000/1.0000/1.0000 & 1.00/1.00/1.00 & 0.9998/1.0000/1.0000 & 1.0000/1.0000/1.0000\\
\rowcolor{gray!15}
T2SMark+P & \textbf{0.00}/\textbf{0.00}/\textbf{0.01} & \textbf{0.4966}/\textbf{0.5037}/\textbf{0.5119} & \textbf{0.5335}/\textbf{0.6868}/\textbf{0.7626    } & \textbf{0.00}/\textbf{0.00}/\textbf{0.01} & \textbf{0.4921}/\textbf{0.5055}/\textbf{0.5116} & \textbf{0.5783}/\textbf{0.6964}/\textbf{0.7551}\\
\bottomrule
\end{tabular}}
\end{table}

%% file: tables/main_results_forgery_imagenet.tex
\begin{table}[t]
\centering
\caption{Results against imprint forgery attack on ImageNet dataset for four semantic watermarking schemes and their PGID-augmented variants. Results are reported for the attack at 50/100/150 optimization steps. Lower AUC and Det. indicate higher robustness to the attack. The best result for each scheme is in \textbf{bold}.}
\vspace{1ex}
\label{tab:forgery_attack_imagenet}
\setlength{\tabcolsep}{1mm}
\resizebox{\textwidth}{!}{
\begin{tabular}{@{}l ccc ccc@{}}
\toprule
\multirow{2}{*}{Method} & \multicolumn{3}{c}{Attacker Model: SD 2.1} & \multicolumn{3}{c}{Attacker Model: SD 1.5} \\
\cmidrule(lr){2-4} \cmidrule(l){5-7}
& Det.$\downarrow$ & Bit Acc.$\downarrow$ & AUC$\downarrow$ & Det.$\downarrow$ & Bit Acc.$\downarrow$ & AUC$\downarrow$ \\
\midrule
TR        & 0.99/1.00/1.00 & --- & 0.9994/0.9998/0.9998 & 0.99/1.00/1.00 & --- & 0.9990/1.0000/1.0000\\
\rowcolor{gray!15}
TR+P      & \textbf{0.06}/\textbf{0.08}/\textbf{0.12} & --- & \textbf{0.5642}/\textbf{0.6324}/\textbf{0.6915} & \textbf{0.05}/\textbf{0.09}/\textbf{0.11} & --- & \textbf{0.5653}/\textbf{0.6295}/\textbf{0.7030}\\
GS        & 1.00/1.00/1.00 & 0.9984/0.9992/0.9996 & 1.0000/1.0000/1.0000 & 1.00/1.00/1.00 & 0.9972/0.9986/0.9991 & 1.0000/1.0000/1.0000\\
\rowcolor{gray!15}
GS+P      & \textbf{0.00}/\textbf{0.04}/\textbf{0.08} & \textbf{0.5479}/\textbf{0.5895}/\textbf{0.6210} & \textbf{0.7801}/\textbf{0.8968}/\textbf{0.9288} & \textbf{0.01}/\textbf{0.06}/\textbf{0.14} & \textbf{0.5512}/\textbf{0.5970}/\textbf{0.6273} & \textbf{0.8044}/\textbf{0.8971}/\textbf{0.9367} \\
PRC       & 0.99/0.99/1.00 & 0.9846/0.9951/0.9948 & --- & 0.89/0.98/0.99 & 0.8815/0.9655/0.9744 & ---\\
\rowcolor{gray!15}
PRC+P     & \textbf{0.00}/\textbf{0.00}/\textbf{0.04} & \textbf{0.5012}/\textbf{0.4967}/\textbf{0.5048} & --- & \textbf{0.00}/\textbf{0.00}/\textbf{0.00} & \textbf{0.4944}/\textbf{0.5038}/\textbf{0.4986} & --- \\
T2SMark   & 1.00/1.00/1.00 & 0.9999/1.0000/1.0000 & 1.0000/1.0000/1.0000 & 1.00/1.00/1.00 & 0.9999/1.0000/1.0000 & 1.0000/1.0000/1.0000\\
\rowcolor{gray!15}
T2SMark+P & \textbf{0.00}/\textbf{0.02}/\textbf{0.06} & \textbf{0.5000}/\textbf{0.5143}/\textbf{0.5250} & \textbf{0.5339}/\textbf{0.6426}/\textbf{0.7452} & \textbf{0.00}/\textbf{0.04}/\textbf{0.07} & \textbf{0.5027}/\textbf{0.5114}/\textbf{0.5423} & \textbf{0.5593}/\textbf{0.6295}/\textbf{0.7030}\\
\bottomrule
\end{tabular}}
\end{table}

%% file: tables/watermarked.tex
\begin{table}[t]
\centering
\caption{Robustness and separability of authentic watermarks after passing through the PGID-F defense. AUC is calculated for the attacks at 50, 100, and 150 steps (50/100/150). Higher Detection Rate, Bit Accuracy, and AUC indicate that the defense preserves the authentic watermark while effectively separating it from forged images.}
\vspace{1ex}
\label{tab:watermarked_robustness}
\setlength{\tabcolsep}{1mm}
\resizebox{\textwidth}{!}{ 
\begin{tabular}{@{}l cc cc cc@{}}
\toprule
\multirow{2}{*}{Method} & \multirow{2}{*}{Det.$\uparrow$} & \multirow{2}{*}{Bit Acc.$\uparrow$} & \multicolumn{2}{c}{AUC$\uparrow$(MS-COCO)} & \multicolumn{2}{c}{AUC$\uparrow$(ImageNet)} \\
\cmidrule(lr){4-5} \cmidrule(l){6-7}
& & & Atk Model: SD 2.1 & Atk Model: SD 1.5 & Atk Model: SD 2.1 & Atk Model: SD 1.5 \\
\midrule 
TR+P      & 0.96 & ---    & 0.9930/0.9902/0.9851 & 0.9930/0.9886/0.9854 & 0.9913/0.9874/0.9805 & 0.9916/0.9883/0.9802 \\
GS+P      & 1.00 & 0.8947 & 1.0000/1.0000/1.0000 & 1.0000/1.0000/1.0000 & 1.0000/0.9996/0.9986 & 1.0000/0.9992/0.9972 \\
PRC+P     & 0.93 & 0.8452 & --- & --- & --- & --- \\
T2SMark+P & 1.00 & 0.9591 & 1.0000/1.0000/1.0000 & 1.0000/1.0000/0.9998 & 1.0000/0.9996/0.9970 & 1.0000/0.9983/0.9940 \\
\bottomrule
\end{tabular}}
\end{table}

%% file: tables/PixArt_both.tex
\begin{table}[t]
\centering
\begin{minipage}{0.49\textwidth}
    \centering
    \caption{Results against imprint removal attack using \textit{PixArt-$\alpha$} as the service provider model.}
    \vspace{1ex}
    \label{tab:pixart_removal_attack}
    \setlength{\tabcolsep}{1.5mm} 
    \resizebox{\textwidth}{!}{ 
    \begin{tabular}{@{}l ccc @{}}
    \toprule
    \multirow{2}{*}{Method} & \multicolumn{3}{c}{Attacker Model: SD 2.1} \\
    \cmidrule(l){2-4}
    & Det.$\uparrow$ & Bit Acc.$\uparrow$ & AUC$\uparrow$ \\
    \midrule
    TR        & 0.20/0.06/0.03 & --- & 0.4170/0.2103/0.1307 \\
    \rowcolor{gray!15}
    TR+P      & \textbf{0.93}/\textbf{1.00}/\textbf{1.00} & --- & \textbf{0.9867}/\textbf{1.0000}/\textbf{0.9978} \\
    GS        & 0.03/0.00/0.00 & 0.4444/0.0927/0.0264 & 0.3608/0.0000/0.0000 \\
    \rowcolor{gray!15}
    GS+P      & \textbf{1.00}/\textbf{1.00}/\textbf{1.00} & \textbf{0.9178}/\textbf{0.9846}/\textbf{0.9932} & \textbf{1.0000}/\textbf{1.0000}/\textbf{1.0000} \\
    PRC       & 0.50/0.00/0.00 & 0.5439/0.5033/0.5013 & --- \\
    \rowcolor{gray!15}
    PRC+P     & \textbf{1.00}/\textbf{0.97}/\textbf{1.00} & \textbf{1.0000}/\textbf{0.7085}/\textbf{0.9475} & --- \\
    T2SMark   & 0.30/0.97/1.00 & 0.5434/0.4932/0.4956 & 0.9020/\textbf{1.0000}/\textbf{1.0000} \\
    \rowcolor{gray!15}
    T2SMark+P & \textbf{0.93}/\textbf{1.00}/\textbf{1.00} & \textbf{0.9298}/\textbf{0.9957}/\textbf{0.9975} & \textbf{0.9933}/\textbf{1.0000}/\textbf{1.0000} \\
    \bottomrule
    \end{tabular}}
\end{minipage}\hfill
\begin{minipage}{0.49\textwidth}
    \centering
    \caption{Results against imprint forgery attack using \textit{PixArt-$\alpha$} as the service provider model.}
    \vspace{1ex}
    \label{tab:pixart_forgery_attack}
    \setlength{\tabcolsep}{1.5mm}
    \resizebox{\textwidth}{!}{
    \begin{tabular}{@{}l ccc @{}}
    \toprule
    \multirow{2}{*}{Method} & \multicolumn{3}{c}{Attacker Model: SD 2.1} \\
    \cmidrule(l){2-4}
    & Det.$\downarrow$ & Bit Acc.$\downarrow$ & AUC$\downarrow$ \\
    \midrule
    TR        & 0.97/0.97/0.97 & --- & 0.9967/0.9993/0.9993 \\
    \rowcolor{gray!15}
    TR+P      & \textbf{0.10}/\textbf{0.17}/\textbf{0.23} & --- & \textbf{0.5178}/\textbf{0.5811}/\textbf{0.6367} \\
    GS        & 1.00/1.00/1.00 & 0.9772/0.9893/0.9943 & 1.0000/1.0000/1.0000 \\
    \rowcolor{gray!15}
    GS+P      & \textbf{0.00}/\textbf{0.07}/\textbf{0.27} & \textbf{0.5715}/\textbf{0.6240}/\textbf{0.6655} & \textbf{0.9322}/\textbf{0.9772}/\textbf{0.9872} \\
    PRC       & 0.97/1.00/1.00 & 0.9858/1.0000/1.0000 & --- \\
    \rowcolor{gray!15}
    PRC+P     & \textbf{0.00}/\textbf{0.00}/\textbf{0.00} & \textbf{0.5055}/\textbf{0.4960}/\textbf{0.4964} & --- \\
    T2SMark   & 1.00/1.00/1.00 & 0.9905/0.9967/0.9979 & 1.0000/1.0000/1.0000 \\
    \rowcolor{gray!15}
    T2SMark+P & \textbf{0.00}/\textbf{0.00}/\textbf{0.13} & \textbf{0.5016}/\textbf{0.5435}/\textbf{0.6301} & \textbf{0.0056}/\textbf{0.2378}/\textbf{0.3833} \\
    \bottomrule
    \end{tabular}}
\end{minipage}
\end{table}

%% file: tables/PixArt_robustness.tex
\begin{wraptable}{r}{0.45\textwidth}
\vspace{-0.5cm}
\centering
\caption{Robustness and separability of authentic watermarks after passing through the PGID-F defense, using \textit{PixArt-$\alpha$} as the service provider model.}
\label{tab:pixart_watermarked_robustness}
\setlength{\tabcolsep}{1.5mm}
\resizebox{\linewidth}{!}{ 
\begin{tabular}{@{}l ccc@{}}
\toprule
\multirow{2}{*}{Method} & \multirow{2}{*}{Det.$\uparrow$} & \multirow{2}{*}{Bit Acc.$\uparrow$} & Attacker: SD 2.1 \\
\cmidrule(l){4-4} 
& & & AUC$\uparrow$ \\
\midrule 
TR+P      & 0.93 & ---    & 0.9822/0.9722/0.9644 \\
GS+P      & 1.00 & 0.9313 & 1.0000/1.0000/0.9978 \\
PRC+P     & 1.00 & 0.6337 & --- \\
T2SMark+P & 0.93 & 0.9780 & 1.0000/0.9922/0.9711 \\
\bottomrule
\end{tabular}}
\end{wraptable}

%% file: sec/6_conclusion.tex
\section{Conclusion}
In this paper, we propose a novel plug-and-play, training-free unified framework capable of protecting semantic watermark methods against both imprint removal and forgery attacks. Based on the observation that imprint attacks maliciously displace the latents out of their respective watermarked or unwatermarked region, our method effectively projects the attacked latents back to their original region. Our experiments demonstrate that PGID effectively restores obscured watermarks against removal attacks and reliably isolates forged images from authentic ones across multiple state-of-the-art watermarking schemes and different model architectures. However, we acknowledge that applying our method to different schemes and architectures might require users to tailor hyperparameters to their specific setup rather than relying on a single default to fully optimize the defense. Despite that, our novel approach opens the path for more robust and reliable content provenance against adversarial threats in watermarking systems.

%% file: sec/7_appendix.tex
\appendix
\addcontentsline{toc}{part}{Appendix}

\section*{Table of Contents: Appendix}

\setcounter{tocdepth}{2} 
\etocsettocstyle{\hypersetup{linkcolor=blue}}{} 

\localtableofcontents 

\newpage 
\newpage
\section{Implementation Details} \label{appendix:implement_details} 
\subsection{Semantic Watermarking}
\label{appendix:semantic_watermark}
\begin{figure}[t]
    \centering
    \includegraphics[width=0.9\linewidth]{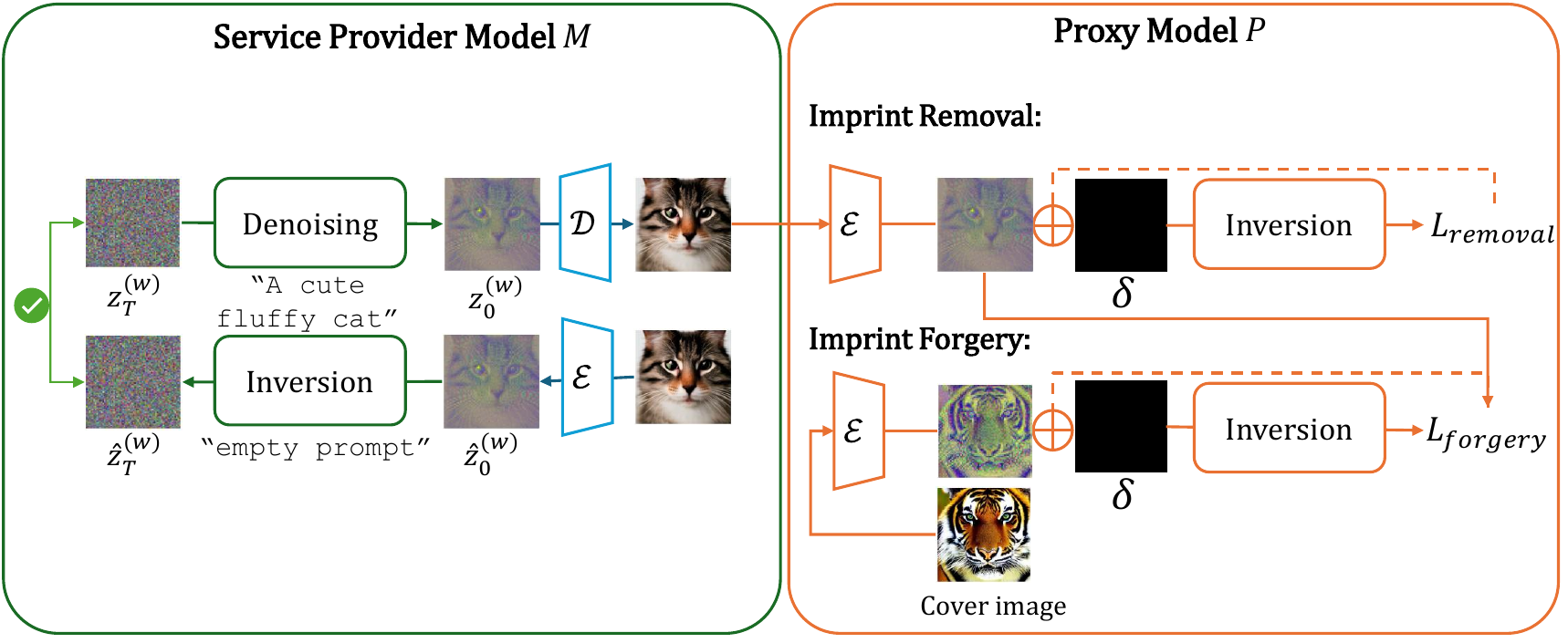}
    \caption{The left side presents the semantic watermarking paradigm. The initial noise contains a watermark pattern, which can be verified after inversion. The right side illustrates imprint removal and forgery attacks. From a proxy diffusion model $P$, the attacks optimize a perturbation and add it to either a watermarked image or a cover image in latent space to yield deceptive results.}
    \label{fig:semantic_imprint}
    \vspace{-0.2cm}
\end{figure}

In this section, we present more details on the evaluated semantic watermarking methods and our implementation settings. Figure~\ref{fig:semantic_imprint} describes the main methodology of semantic watermarks and imprint attacks.

\textbf{Tree Ring. }Tree-Ring watermarking \cite{wen2023treering} is the pioneering work of the semantic watermark paradigm. It embeds a structured pattern into the Fourier space of the initial noise vector used during the diffusion sampling process, allowing the watermark to be robustly detected by inverting the generated image back to its initial latent noise state. After obtaining the estimated initial noise, the method calculates a test statistic by summing the squared absolute differences between the observed and expected frequency values within the embedded rings. Under the null hypothesis ($H_0$) that an unwatermarked image's inverted initial noise follows a Gaussian distribution with an unknown variance, the test statistic conforms to a noncentral $\chi^2$ distribution, enabling the watermarker to calculate a $p$-value. If it falls below a predefined significance threshold $\tau$, $H_0$ is rejected, determining the image as watermarked.

In our experiments, we adhere to the original paper by embedding a ring pattern using a circular mask with a radius of 10 and employing DDIM as the scheduler. As the threshold is not reported in the original paper, for our experiments, we compute the threshold $\tau$ from 5,000 generated unwatermarked images for the desired FPR. The thresholds are set at $0.05385$ and $0.04743$, corresponding to SD 2.1 and PixArt-$\alpha$ for FPR = $1\%$, respectively. For AUC calculation, we use the $p$-value as the continuous score.

\textbf{Gaussian Shading. }Unlike Tree-Ring, which causes the initial noise to deviate from a Gaussian distribution, resulting in performance loss, Gaussian Shading~\cite{yang2024gaussian} embeds watermark information while preserving the noise distribution. Here we focus on the standard setting of $l=1$ from the original paper. In this setting, an encrypted message $m$ derived from the original bit message $s$ determines the signs of the initial noise. The watermark is detected by inverting the image, decrypting the retrieved message $s'$, and comparing it to reference messages.

Under the null hypothesis of an unwatermarked image, the extracted bits follow a Bernoulli distribution with parameter $p = 0.5$. When comparing $s'$ to a single reference message of length $k$, the integer count of matching bits, denoted as $c$, follows a binomial distribution $B(k, 0.5)$. The probability of observing more than $c_{\tau}$ matching bits by chance is given by the regularized incomplete beta function:
\begin{equation}
    \text{FPR}(c_{\tau}) = P(c > c_{\tau}) = B_{1/2}(c_{\tau} + 1, k - c_{\tau})
\end{equation}

In zero-bit watermarking, $s'$ is compared to a single predefined message, bounded directly by $\text{FPR}(c_{\tau})$. However, in multi-bit watermarking, $s'$ is compared against $N$ possible user IDs known to the service provider, and the match with the highest accuracy is selected. Because this constitutes a multiple testing problem, the overall False Positive Rate is adjusted depending on the number of users $N$:
\begin{equation}
    \text{FPR}(c_{\tau}, N) = 1 - (1 - \text{FPR}(c_{\tau}))^N
\end{equation}

For evaluation, we follow Müller et al. (2025)~\cite{Muller_2025_CVPR} and map the integer count threshold $c_{\tau}$ to a bit accuracy threshold, $\tau = \frac{c_{\tau}}{k}$. In our experiments, we focus on the multi-watermarking scenario with $N=100,000$ as in~\cite{yang2024gaussian} and target $\text{FPR}(c_{\tau}, N) = 10^{-6}$, which yields $\tau = 0.70703$~\cite{Muller_2025_CVPR}. We use the DDIM sampler and parameter settings from Yang et al. (2024)\cite{yang2024gaussian} with $k = 256$, $\rho = 64$, and $l = 1$. For AUC calculation, we use the bit accuracy as the continuous score.

\textbf{PRC. } The PRC watermark builds upon Gaussian Shading by introducing a pseudorandom error-correcting code (PRC)~\cite{christ2024pseudorandomerrorcorrectingcodes} for better undetectability. This method embeds a message by sampling the initial noise vector directly from a cryptographically secure codebook, ensuring the watermarked latents are computationally indistinguishable from standard Gaussian noise. Detection follows the standard inversion paradigm to recover the estimated initial noise. However, unlike previous methods that use a globally fixed threshold, the PRC watermark utilizes an image-dependent detection logic based on soft-decision decoding. Specifically, the decoding process assigns confidence weights to the estimated initial noise values based on their magnitudes, treating larger values as more reliable indicators of the watermark. Based on these weights, a dynamic threshold $\tau$ is calculated for each image using Hoeffding's inequality to strictly bound the False Positive Rate (FPR).

Following the settings in Gunn et al. (2025)\cite{gunn2025undetectable}, we implement PRC using the DPM scheduler for generation and the exact inversion method from Hong et al. (2023)\cite{hong2023exact} for inversion. Because the dynamically calculated threshold varies across images, there is no globally comparable continuous test statistic. Consequently, we omit the AUC for the PRC scheme and instead evaluate its performance using the True Positive Rate (TPR) at a targeted FPR of $10^{-6}$.

\textbf{T2SMark. } T2SMark focuses on resolving the fundamental trade-off between watermark robustness and generation diversity. To achieve this, it introduces Tail-Truncated Sampling (TTS), which selectively embeds watermark information exclusively into the more reliable tail regions of the Gaussian distribution. The central region is left for random sampling to maintain the overall noise distribution and preserve sample diversity. Furthermore, it employs a two-stage hierarchical key encryption framework to randomize the codewords. For detection, following the standard inversion process, the estimated initial noise vector is projected onto the normal vectors of the first stage. The detection test statistic is computed as the $L_1$ norm of this first-stage projection vector. Under the null hypothesis of an unwatermarked image, this statistic is compared against a threshold $d$ that is calibrated to a target False Positive Rate (FPR). If the test statistic exceeds $d$, the image is determined to be watermarked. 

In our experiments, we follow the default settings from the original paper, including the use of a DDIM scheduler and the empirically optimal truncation threshold $\tau = 0.674$. For detection evaluation, the threshold $d$ is calibrated to achieve a target FPR of $10^{-6}$ to remain consistent with the original work. Since the original paper does not report the threshold $d$, we empirically estimate the null distribution using 10,000 unwatermarked images. By assuming normality and computing the critical value at the $10^{-6}$ upper tail, we obtain $d=260.7018$ and $d=250.6571$ for SD 2.1 and PixArt-$\alpha$, respectively. We use the test statistic, which is the $L_1$ norm of the first-stage projection vector, for the calculation of the AUC.

\subsection{Imprint Attacks}
\textbf{Imprint Removal Attack. }Proposed by Müller et al. (2025)\cite{Muller_2025_CVPR}, the attack aims to remove the watermark from a watermarked image $x^{(w)}$ by using a proxy model and optimizing a perturbation $\delta$ in the latent space with the following loss function:
\begin{equation} 
    \mathcal{L}_{\text{removal}} = \| I_{0\to T} (\hat{z}_{0}^{(w)} + \delta) + \hat{z}_{T}^{(w)} \|_2 
\end{equation}
where $I_{0\to T}(\cdot)$ is the diffusion inversion process using the proxy model, $\hat{z}_{T}^{(w)}$ is the estimated initial noise from the watermarked image, and $\hat{z}_{0}^{(w)} = \mathcal{E}(x^{(w)})$ with $\mathcal{E}$ being the encoder from the proxy model. The attack pushes the estimated initial noise away from its original value. The final image with the watermark removed is obtained by decoding the perturbed latent: $x^{(w)}_{removed} = \mathcal{D}(\hat{z}_{0}^{(w)}+\delta)$.

\textbf{Imprint Forgery Attack. }On the other hand, the imprint forgery attack aims to imprint the watermark from a watermarked image $x^{(w)}$ to a cover image $x^{(c)}$ by optimizing a perturbation $\delta$ in the latent space with the following loss function:
\begin{equation} 
    \mathcal{L}_{\text{forgery}} = \| I_{0\to T} (\hat{z}_{0}^{(c)} + \delta) - \hat{z}_{T}^{(w)} \|_2 
\end{equation}
where $I_{0\to T}(\cdot)$ is the diffusion inversion process using the proxy model, $\hat{z}_{T}^{(w)}$ is the estimated initial noise from the watermarked image, and $\hat{z}_{0}^{(c)} = \mathcal{E}(x^{(c)})$ with $\mathcal{E}$ being the encoder from the proxy model. Opposite to the removal attack, the introduced perturbation guides the estimated initial noise of the cover image towards the watermarked image, thereby fooling the watermark detector. The final forged image is obtained by decoding the perturbed latent: $x^{(c)}_{forged} = \mathcal{D}(\hat{z}_{0}^{(c)} + \delta)$.

Following Müller et al. (2025)\cite{Muller_2025_CVPR}, we report the results on 50, 100, and 150 optimization steps with a learning rate of 0.01. In our main experiments, we utilize DDIM as the attacker's scheduler and consider both white-box and grey-box scenarios where the proxy model used by the attacker is either the same as or different from the service provider model.
\subsection{Hyperparameter settings:}\label{appendix:hypersettings_in_paper}
\textbf{SD 2.1 (Main experiments). }With SD 2.1 as the service provider model, we follow the settings:
\begin{itemize}
    \item PGID-R (Against removal): We set the stopping timestep $k=10$, number of skip inversions $s=1$, and guidance strength $\gamma = 0.045$ for all watermarking schemes. 
    \item PGID-F (Against forgery): We set the stopping timestep $k=15$, number of skip inversions $s=3$, and guidance strength $\gamma = 0.001$ for Tree Ring, Gaussian Shading, and T2SMark. We set the stopping timestep $k=7$, number of skip inversions $s=3$, and guidance strength $\gamma = 0.001$ for PRC.
\end{itemize}
\textbf{PixArt-$\alpha$. }With PixArt-$\alpha$ as the service provider model, we report the results with the settings:
    \begin{itemize}
        \item PGID-R (Against removal): We set the stopping timestep $k=10$, number of skip inversions $s=1$, and guidance strength $\gamma = 0.045$ for all watermarking schemes. An exception is that for PRC, we evaluate the attack at $50$ optimization steps with the setting: $k=10$, $s=1$, and $\gamma=0.01$. We discuss the rationale for this specialized configuration in Section~\ref{appendix:guidance_strength}.
        \item PGID-F (Against forgery): We set the stopping timestep $k=8$, number of skip inversions $s=3$, and guidance strength $\gamma = 0.005$ for all schemes.
    \end{itemize}

\section{Ablation Study of Hyperparameter Settings}
\label{appendix:hyperparam_settings} 
\subsection{Number of skip inversions $s$}
In this section, we justify the necessity of asymmetric inversion-denoising cycles and show the effects of the number of skip inversions $s$. We evaluate the impact of the skip parameter $s$ using the following experiment configuration: SD 2.1 for both the service provider and the attacker, with imprint attacks optimized over 150 steps. We set the stopping timestep $k$ and guidance strength $\gamma$ to the values as in our main experiments. Keeping the stopping timestep $k$ and guidance strength $\gamma$ constant, we vary $s\in\{0,1,3,5\}$ to quantify how cycle asymmetry affects performance. 

\begin{figure}[t]
    \centering
    \subcaptionbox{\label{fig:detection_vs_s}}%
    {\includegraphics[width=0.4\linewidth]{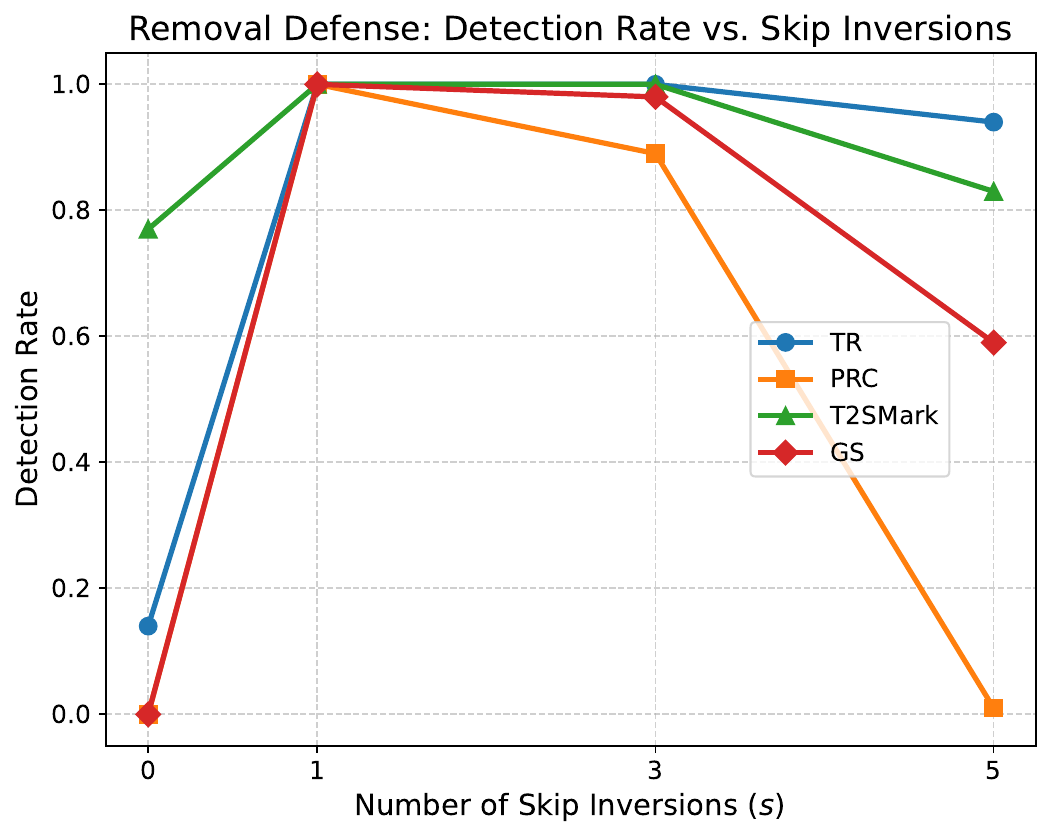}}
    \hspace{0.5cm} 
    \subcaptionbox{\label{fig:mse_vs_s}}%
    {\includegraphics[width=0.4\linewidth]{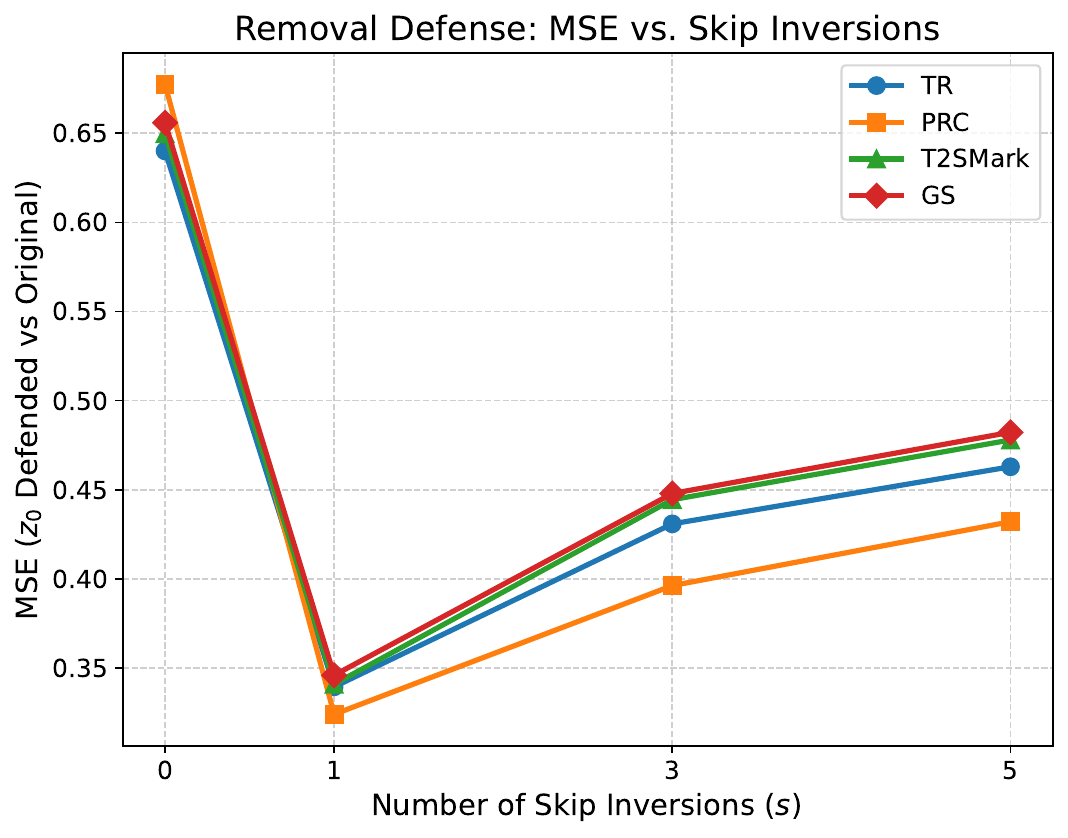}}
    \caption{(a) Detection rate of removal-attacked images under PGID-R with different numbers of skip inversions $s$. (b) MSE between the watermarked latent and the projected latent of watermarking schemes under PGID-R with different numbers of skip inversions. The lowest is achieved at $s=1$, indicating that at this setting, the removal-attacked latent is the closest to the watermarked latent after projection.} 
    \label{fig:combined_vs_s}
\end{figure}

\textbf{Against removal attack. }As described in Figure~\ref{fig:combined_vs_s}, symmetric inversion-denoising cycles ($s=0$) provide nearly no defense for Tree Ring and Gaussian Shading. The setting where $s=1$ achieves the highest detection rate and also the lowest MSE, confirming the need to denoise more steps than inverting. We attribute this improvement to the perturbed nature of the attacked latent. The introduced perturbation essentially acts as a layer of noise on the original latent. Therefore, extra denoising steps leverage the model's natural ability to remove noise, effectively filtering out the adversarial perturbations and pulling the displaced latent back toward the clean latent manifold.
However, with higher $s$, the restoration capability of PGID could potentially become worse. This occurs because excessive denoising over-smooths the latent and causes the model's natural projection to strip away the embedded watermark signal. From Figure~\ref{fig:detection_vs_s}, with $s=3$, the performance for PRC starts to degrade, while other schemes remain nearly unchanged. Only at $s=5$, the performance starts to degrade for other schemes. This can be explained by the sensitivity of each watermarking scheme. PRC employs a more intricate detection mechanism than other schemes, which effectively narrows its watermarked region for a given signal. In other words, PRC is more unforgiving, where a slight difference in the projected latent can push it outside of the watermarked region. Generally, no schemes are affected significantly if $s$ stays close to $1$.

\begin{figure}[t]
    \centering
    \subcaptionbox{\label{fig:detection_vs_s_forgery}}%
    {\includegraphics[width=0.4\linewidth]{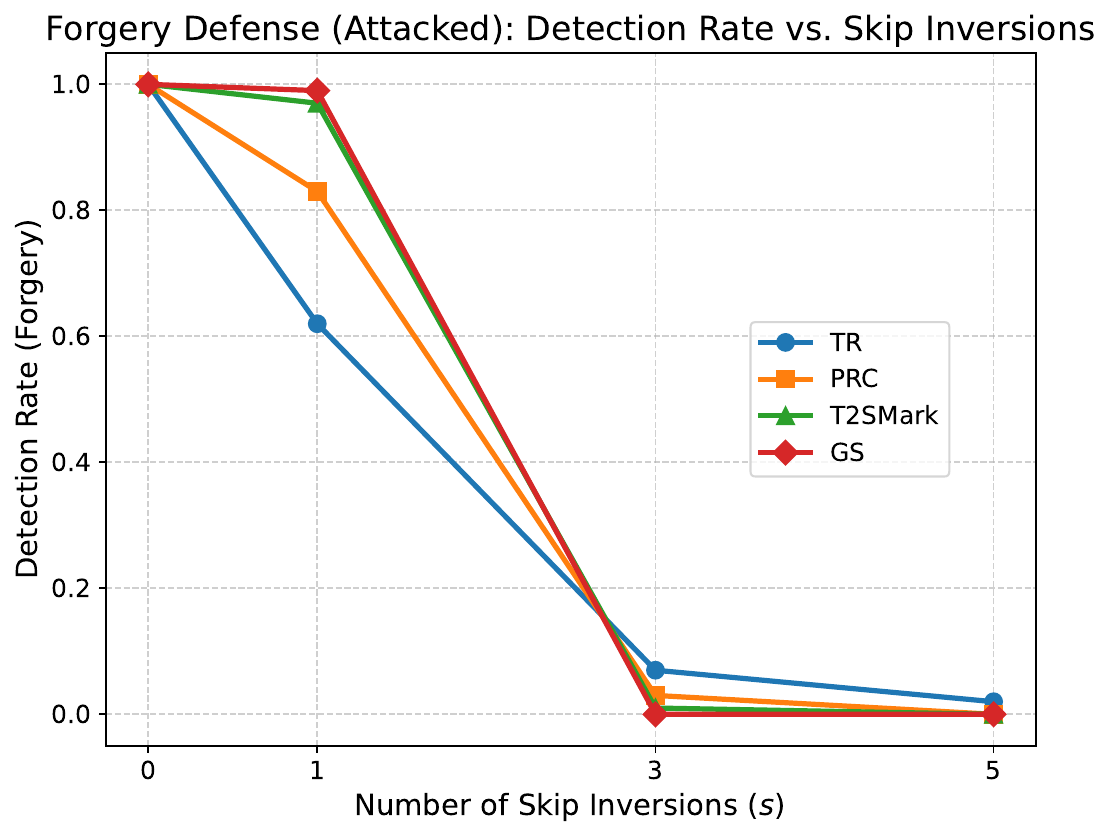}}
    \hspace{0.5cm} 
    \subcaptionbox{\label{fig:mse_vs_s_forgery}}%
    {\includegraphics[width=0.4\linewidth]{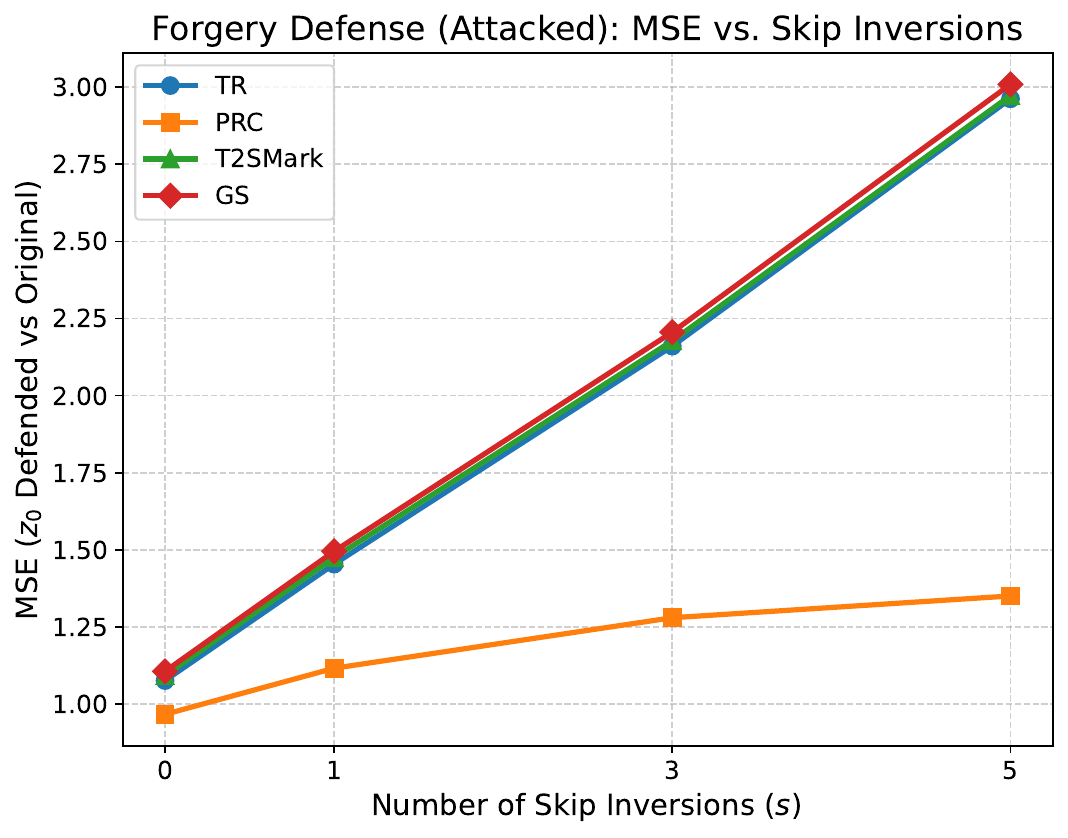}}
    \caption{(a) Detection rate of forged images under PGID-F with different numbers of skip inversions $s$. (b) MSE between watermarked latent and projected latent under PGID-F with different numbers of skip inversions. Initially, forged latents have a small MSE from the watermarked latent. After projection, a higher MSE indicates that the projected latent is pushed away from the watermarked latent, mitigating the forged signals.} 
    \label{fig:combined_vs_s_forgery}
    \vspace{-1.1cm}
\end{figure}

\begin{figure}[t]
    \centering
    \subcaptionbox{\label{fig:detection_vs_s_watermarked}}%
    {\includegraphics[width=0.4\linewidth]{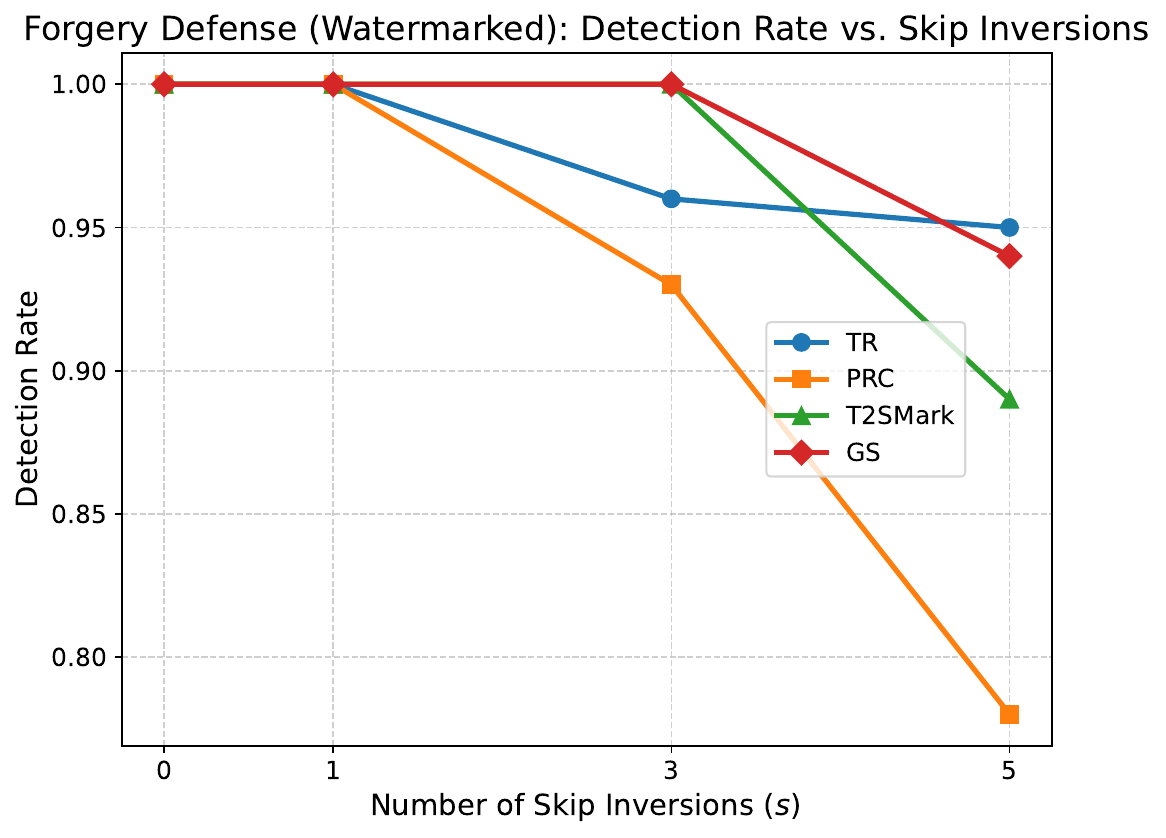}}
    \hspace{0.5cm} 
    \subcaptionbox{\label{fig:mse_vs_s_watermarked}}%
    {\includegraphics[width=0.4\linewidth]{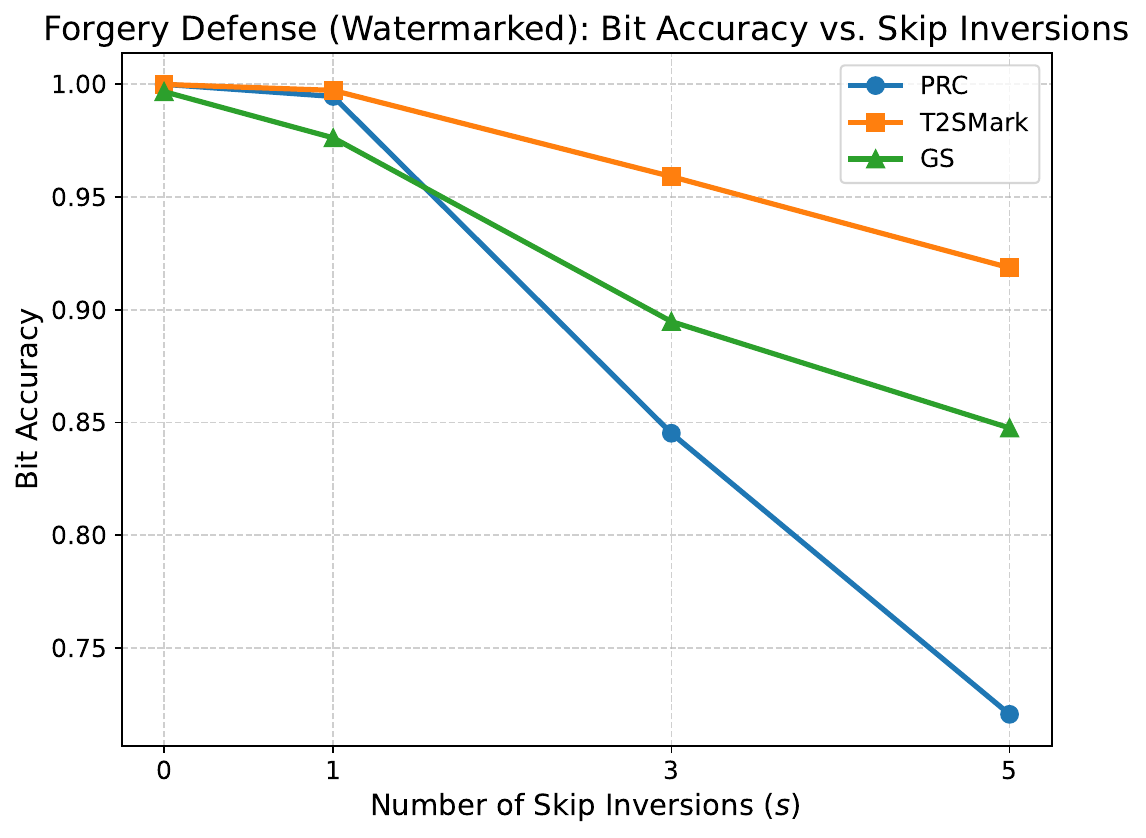}}
    \caption{(a) Detection rate of watermarked images under PGID-F with different numbers of skip inversions $s$. (b) Bit accuracy of watermarked images under PGID-F with different numbers of skip inversions. With a high value of $s$, watermarks from schemes that are not very robust (Tree Ring, PRC) could be falsely denied.} 
    \label{fig:combined_vs_s_watermarked}
\end{figure}

\textbf{Against forgery attack. }From Figure~\ref{fig:combined_vs_s_forgery}, similar to the removal case, setting $s=0$ yields no defense for all schemes (the detection rate remains $100\%$). Our results also show that defense against forgery requires more aggressive latent projection, where defense becomes effective only from $s=3$ with the current stopping timestep $k$ setting. At $s=3$, most forged watermarks are rejected, and the high MSE between the watermarked latent and the projected latent reflects this. However, setting $s$ too high comes at the cost of potentially rejecting authentic watermarks as well if the threshold is set at a very low FPR. As shown in Figure~\ref{fig:combined_vs_s_watermarked}, even though semantic watermarks are inherently robust to the process, some schemes are less robust than others. From the detection rate and bit accuracy of watermarked images, we found PRC and Tree Ring to be the least robust, while Gaussian Shading and T2SMark are the most robust. This claim is supported in Yang et al. (2025)\cite{yang2025t2smark}, where they also showed that PRC and Tree Ring can easily lose their watermark from basic image transformations. 

\subsection{Stopping Timestep $k$}
With the same experiment configuration, in this part, we analyse the effects of the stopping timestep $k$ by fixing the number of inversions $s$ and guidance strength $\gamma$. We vary $k\in\{5,10,15\}$ for removal case and $k\in\{7,10,15\}$ in forgery case. 

\begin{figure}[t]
    \centering
    \includegraphics[width=0.4\linewidth]{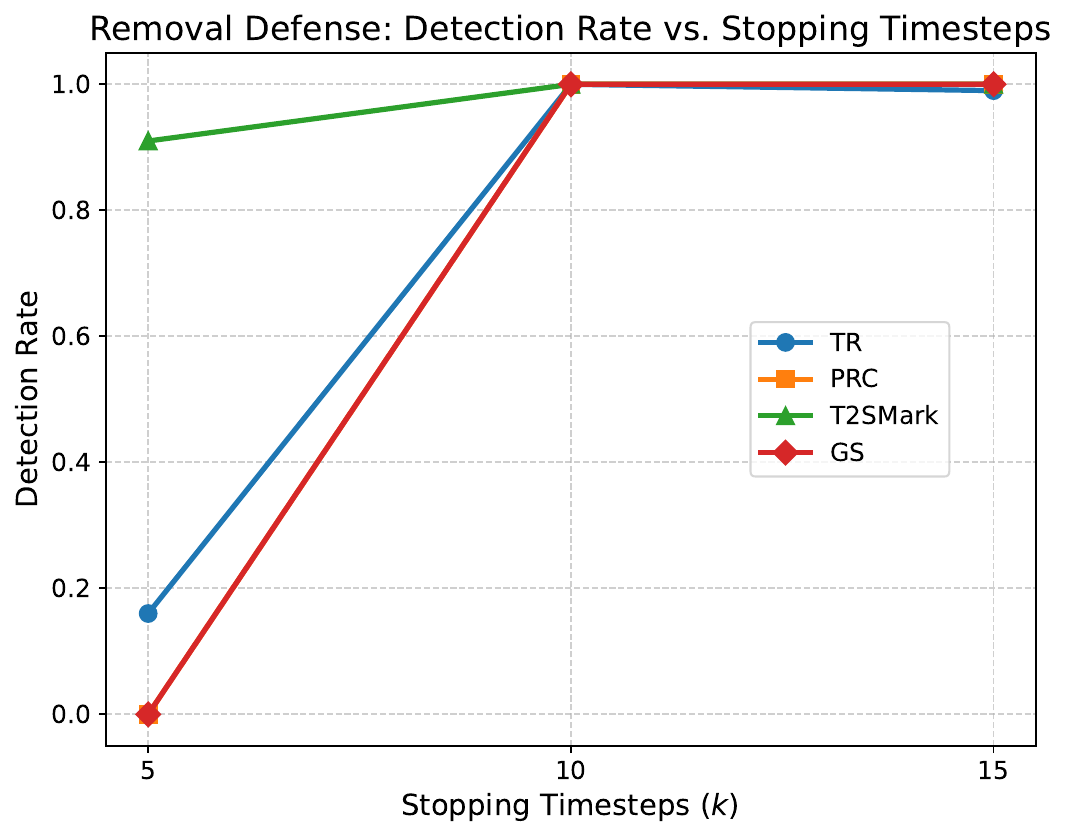}
    \caption{Detection rate of removal-attacked images under PGID-R with different stopping timesteps $k$. With $k$ too small, the latent projection fails to provide good results. However, with sufficiently large $k$, nearly perfect detection rate is achieved.} 
    \label{fig:detection_vs_k}
    \vspace{-0.5cm}
\end{figure}
\textbf{Against removal attack. }Figure \ref{fig:detection_vs_k} illustrates the detection rate of images subjected to removal attack under different stopping timesteps $k$. At $k=5$, most schemes remain unable to detect watermarks. This is because when $k$ is too small, the latent refinement process is confined to the initial timesteps. Consequently, it lacks the necessary leverage to correct the overall inversion trajectory through to the final timestep $T$. Figure~\ref{fig:detection_vs_k} indicates that a near-perfect detection rate is achieved by both $k=10$ and $k=15$. This shows that PGID-R can restore removed watermark signals with a sufficiently large $k$.

\begin{figure}[t]
    \centering
    \subcaptionbox{\label{fig:detection_vs_k_forgery}}%
    {\includegraphics[width=0.4\linewidth]{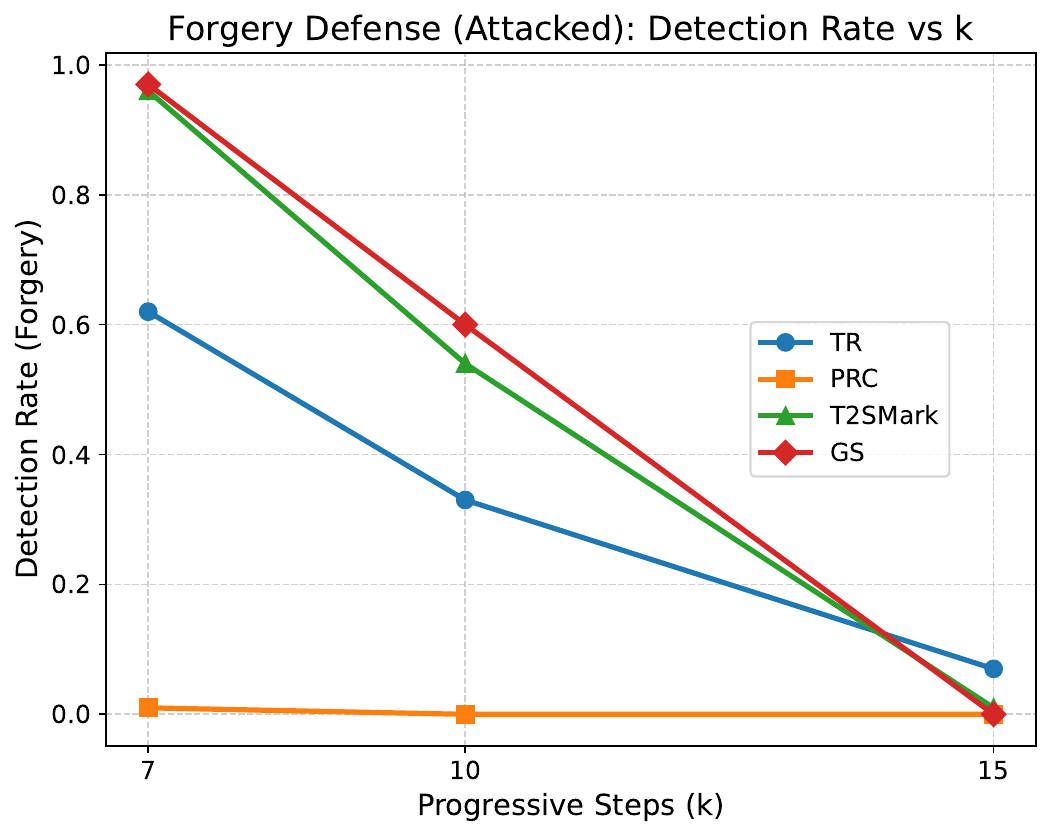}}
    \hspace{0.5cm} 
    \subcaptionbox{\label{fig:detection_vs_k_watermarked}}%
    {\includegraphics[width=0.4\linewidth]{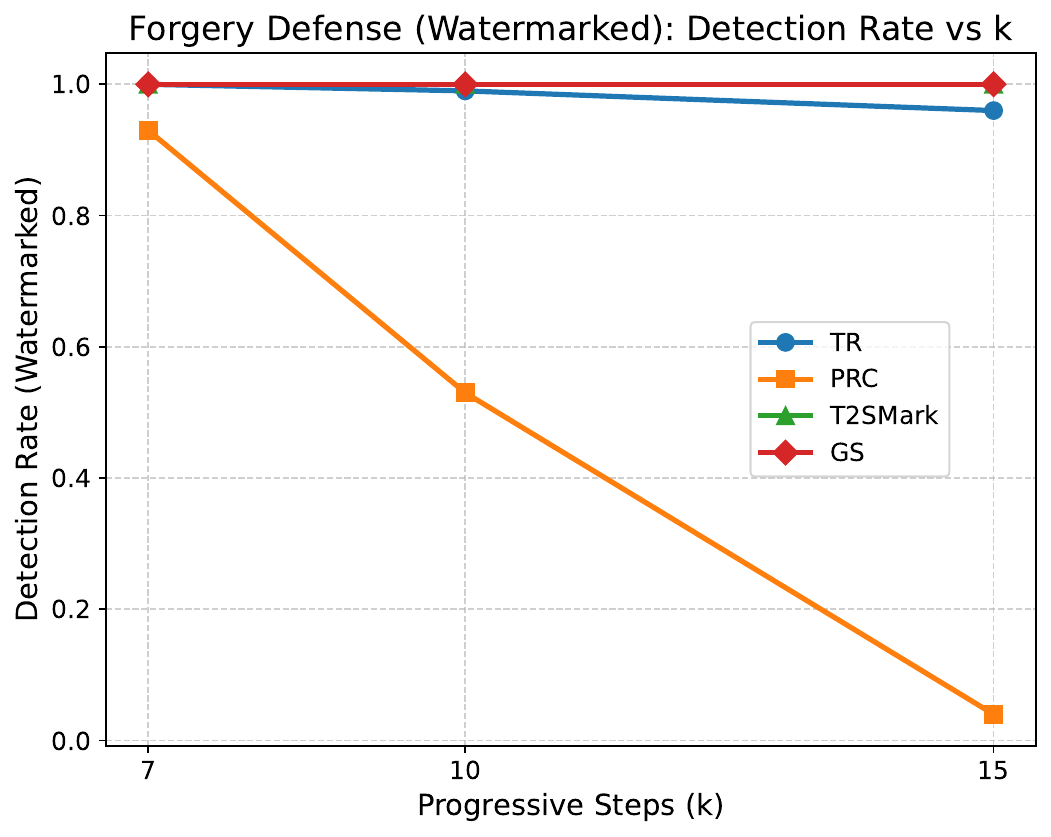}}
    \caption{(a) Detection rate of forged images under PGID-F with different stopping timesteps $k$. (b) Detection rate of watermarked images under PGID-F with different stopping timesteps. With a higher $k$, more forged instances are rejected. However, with a low FPR threshold, authentic watermarks could also be rejected if $k$ is too high. Therefore, $k$ should be selected based on this trade-off.} 
    \label{fig:combined_vs_k_watermarked}
\end{figure}

\textbf{Against forgery attack. }From Figure~\ref{fig:detection_vs_k_forgery}, it can be observed that with $k$ large enough, most of the forged watermark instances are no longer detectable. This can easily be explained similarly to the removal case: with higher $k$, the latent refinement process has more leverage to correct the inversion trajectory. Most forged instances from Gaussian Shading, Tree Ring, and T2SMark are projected back to the unwatermarked region at $k=15$. PRC is a special case where forged images lose all of their signal with only $k=7$. As described in Figure~\ref{fig:detection_vs_k_watermarked}, other schemes maintain a high detection rate at higher $k$, while for PRC, the detection rate degrades heavily. Again, we attribute this to the inherent low robustness of PRC, which makes even forged watermarks easy to remove. As a result, a user should consider a suitable choice of $k$ depending on the robustness of the scheme they are using. Setting $k$ can be thought of as a trade-off: a higher $k$ facilitates rejection of forged instances, but could also potentially reject some authentic watermarked instances at a low FPR threshold setting.

\subsection{Guidance Strength $\gamma$}\label{appendix:guidance_strength}
To better understand the role of the guidance strength $\gamma$ in our refinement process, with the same experiment configuration, we fix the stopping timestep $k$ and the number of inversions $s$ while varying the value of $\gamma $. We consider the values $\gamma\in \{0,0.045,0.075, 0.1\}$ in removal defense, and $\gamma \in \{0,0.001,0.005, 0.01\}$ in forgery defense.
$\gamma = 0$ is equivalent to the scenario where no guidance is introduced. 

\begin{figure}[t]
    \centering
    \subcaptionbox{\label{fig:dectection_vs_guidance}}%
    {\includegraphics[width=0.4\linewidth]{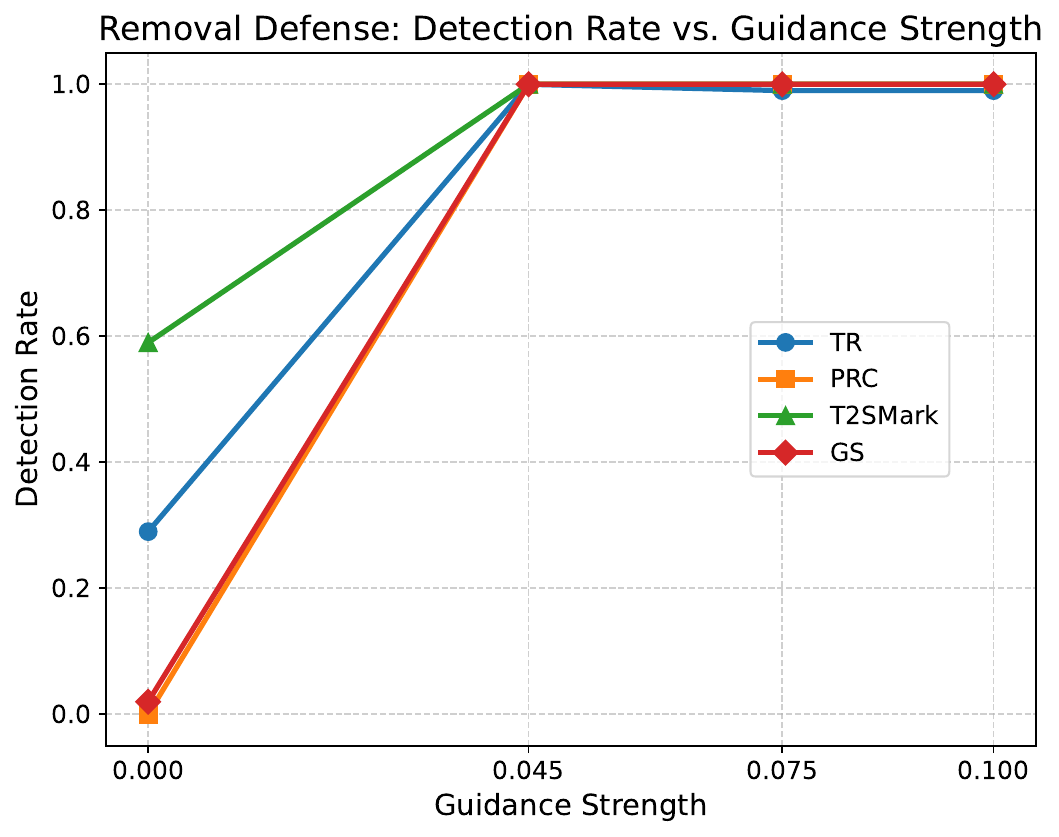}}
    \hspace{0.5cm} 
    \subcaptionbox{\label{fig:bit_vs_guidance}}%
    {\includegraphics[width=0.4\linewidth]{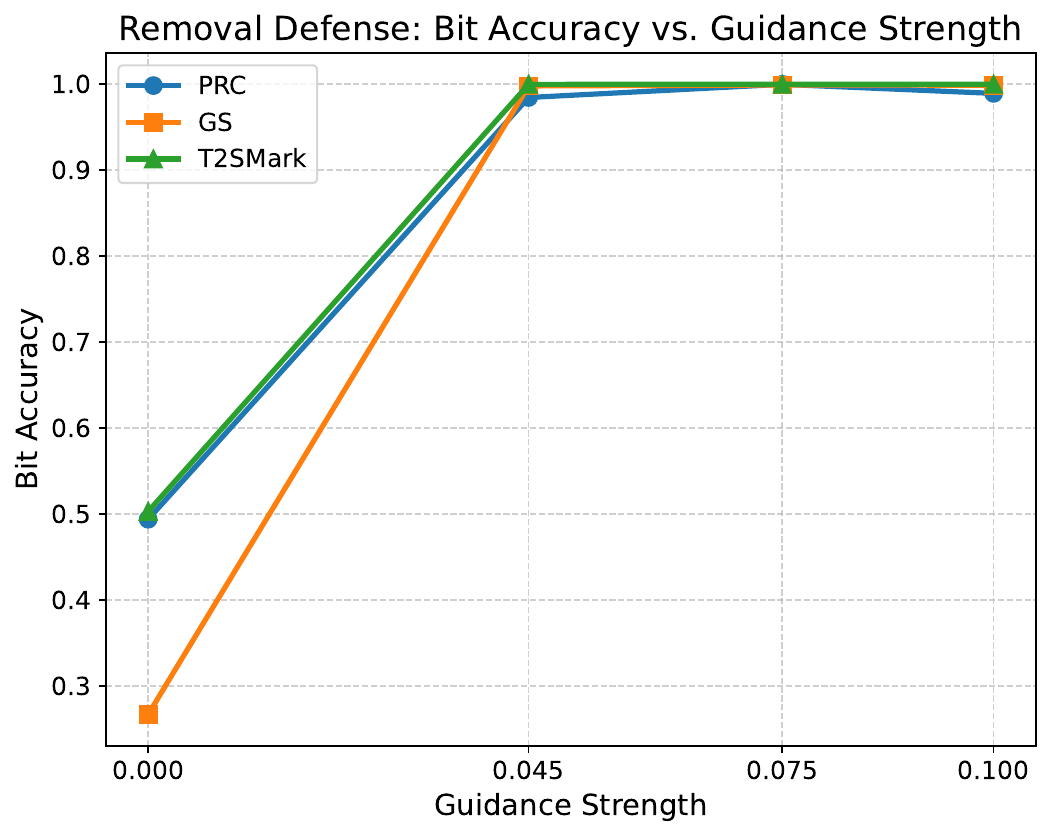}}
    \caption{(a) Detection rate of removal-attacked images under PGID-R with different guidance strength $\gamma$. (b) Bit accuracy of removal-attacked images under PGID-R with different guidance strength $\gamma$. With guidance, the performance is significantly better, highlighting the need for this mechanism. The guidance strength $\gamma$ can also be flexible, where multiple values could provide similarly good results.} 
    \label{fig:combined_vs_guidance}
\end{figure}

\begin{figure}[t]
    \centering
    \subcaptionbox{\label{fig:detection_vs_prc}}%
    {\includegraphics[width=0.4\linewidth]{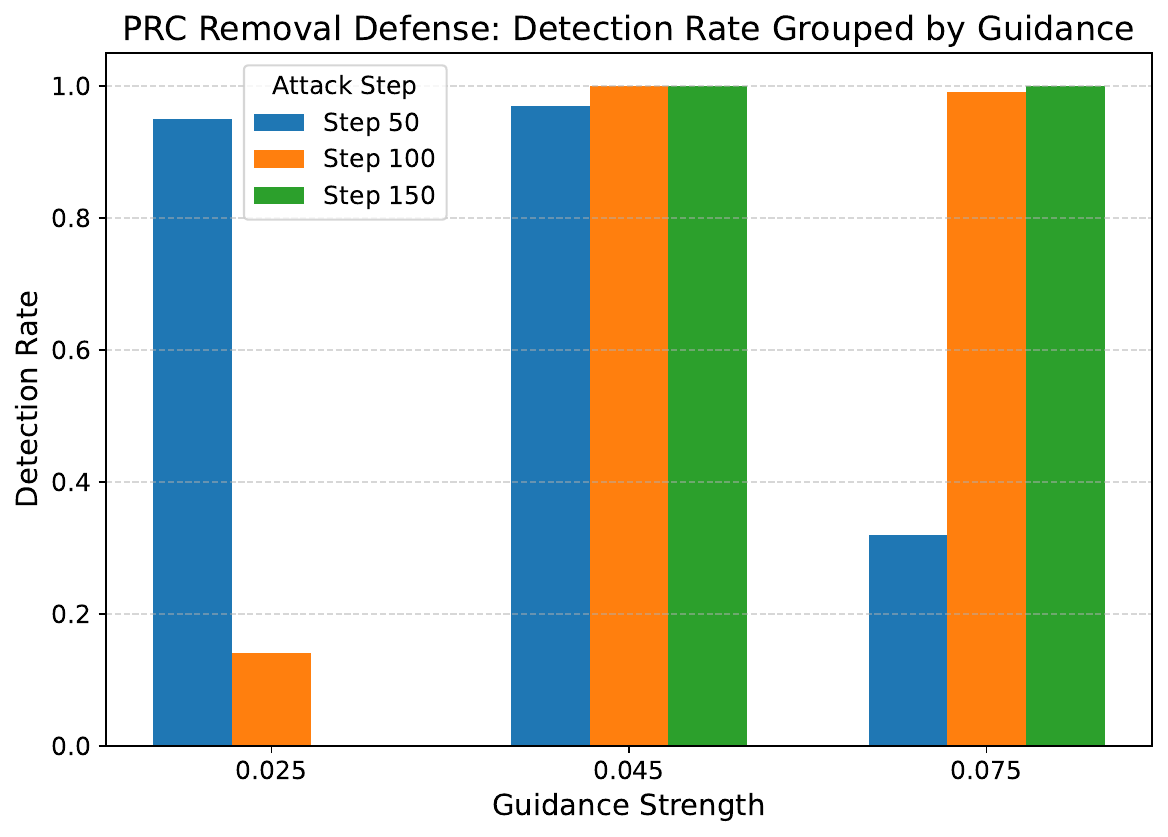}}
    \hspace{0.5cm} 
    \subcaptionbox{\label{fig:bit_accuracy_vs_prc}}%
    {\includegraphics[width=0.4\linewidth]{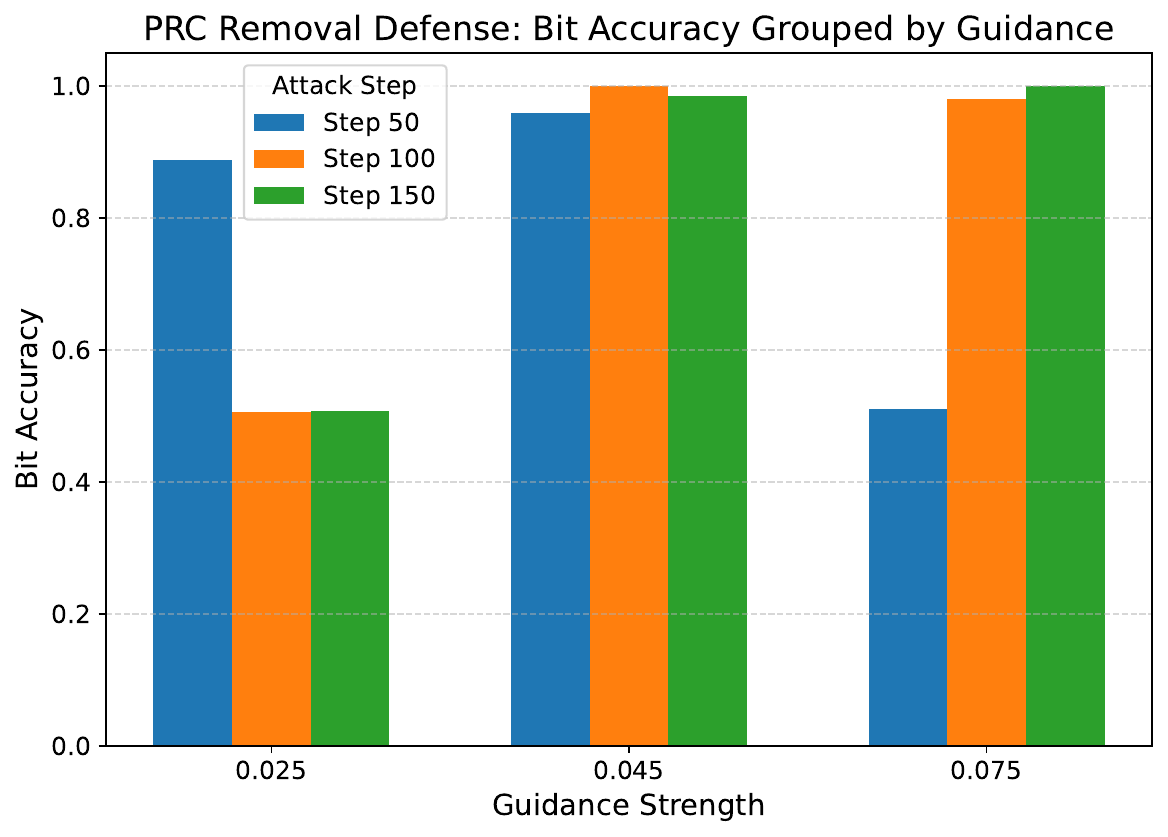}}
    \caption{(a) Detection rate of removal-attacked images at 50, 100, and 150 optimization steps on PRC under PGID-R with different guidance strength $\gamma$. (b) Bit accuracy of removal-attacked images at 50, 100, and 150 optimization steps on PRC under PGID-R with different guidance strength $\gamma$. In the case of PRC, there is a trade-off between protection against the attack at lower and higher optimization steps, where a lower $\gamma$ is better against lower steps, and a higher $\gamma$ is better against higher steps.} 
    \label{fig:combined_vs_prc}
    \vspace{-0.5cm}
\end{figure}

\textbf{Against removal attack. }Figure~\ref{fig:combined_vs_guidance} indicates that the introduced guidance provides superior performance compared to the configuration without guidance. The choice of $\gamma$ could also be relatively versatile, where a higher $\gamma$ yields similar results to the setting in the main experiments ($\gamma=0.045$). However, for schemes with very high sensitivity, such as PRC, we witness a trade-off between the protection against the attack at lower and higher optimization steps. As described in Figure~\ref{fig:combined_vs_prc}, with smaller values of $\gamma$ ($\gamma=0.025$), it might defend against the attack at $50$ steps, but is unable to retrieve the watermarks against an attack with higher steps. 
On the other hand, with larger values of $\gamma$ ($\gamma = 0.075$), only attacks with larger optimization steps can be defended. This is a phenomenon we only observe with PRC. We believe that the high sensitivity of PRC demands a much more precise latent projection compared to other schemes. Consequently, securing such sensitive schemes against varying attack intensities either requires extensive hyperparameter tuning to find a globally optimal $\gamma$, or leveraging multiple parallel PGID instances configured with diverse settings to ensure comprehensive protection. In our experiments with PixArt-$\alpha$, we follow the second approach.

\begin{figure}[t]
    \centering
    \subcaptionbox{\label{fig:dectection_forgery_vs_guidance}}%
    {\includegraphics[width=0.4\linewidth]{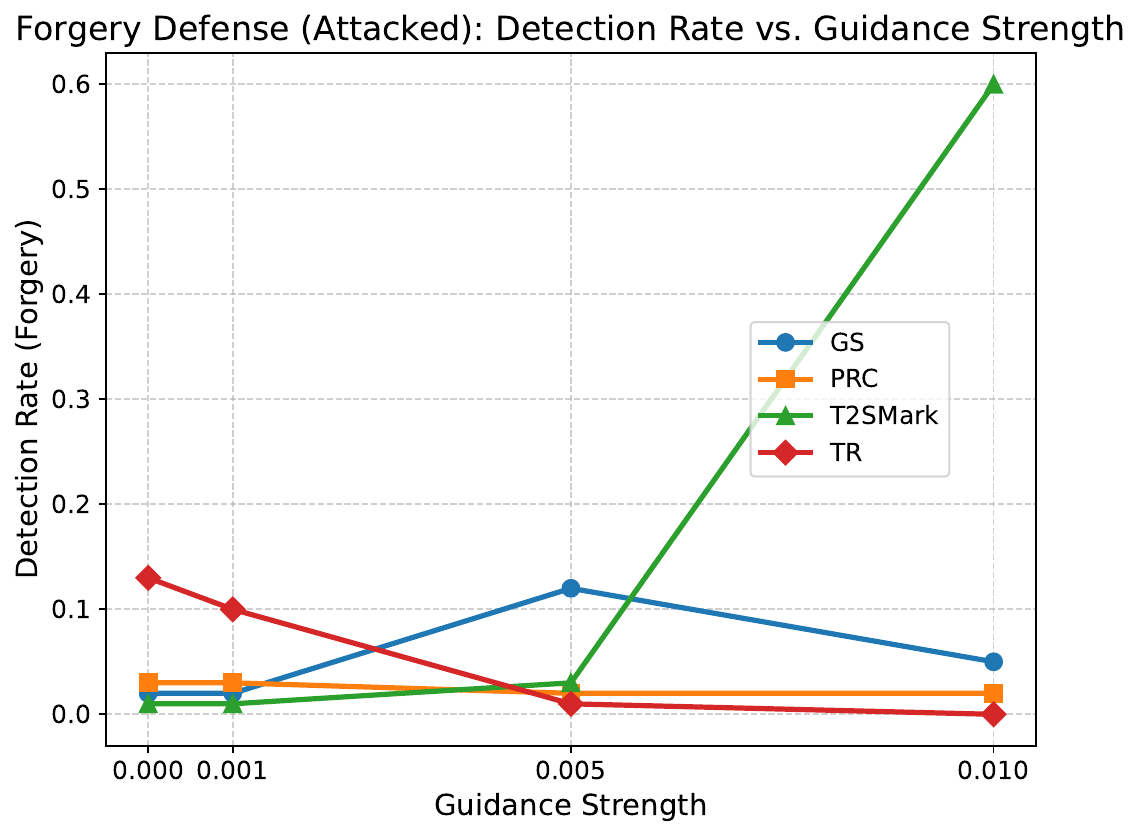}}
    \hspace{0.5cm} 
    \subcaptionbox{\label{fig:bit_forgery_vs_guidance}}%
    {\includegraphics[width=0.4\linewidth]{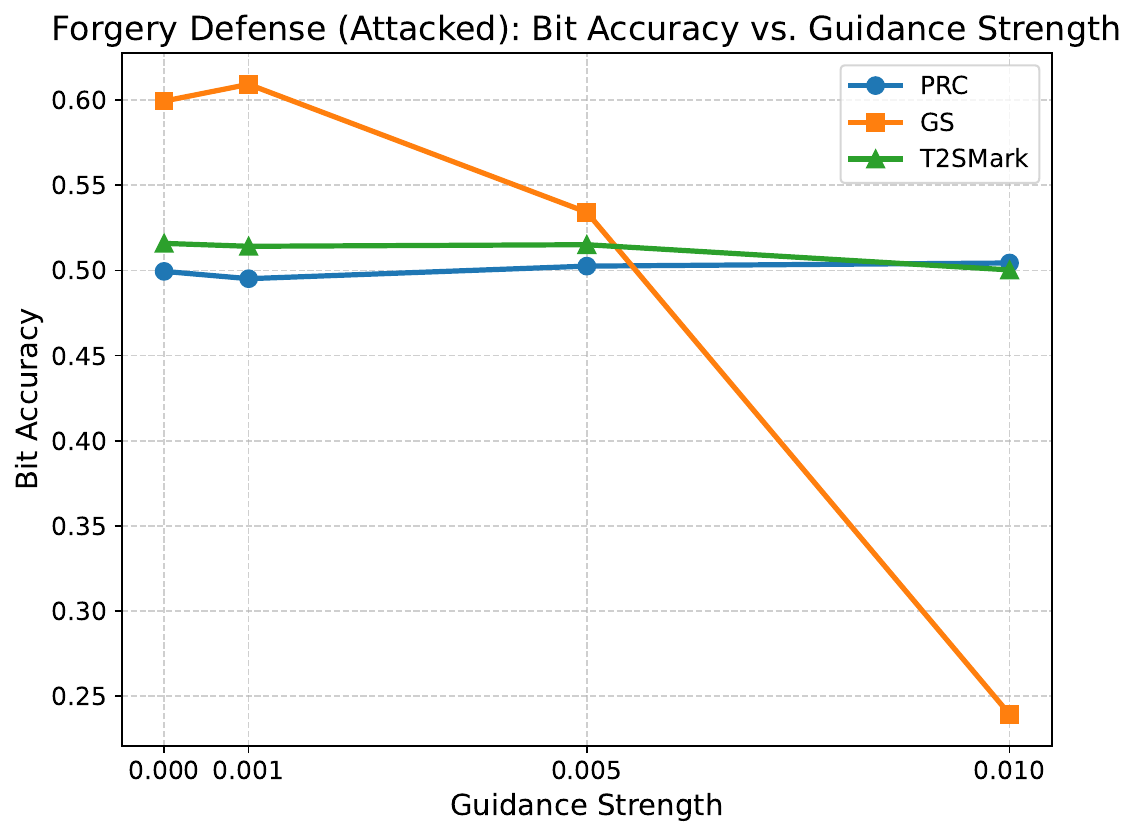}}
    \caption{(a) Detection rate of forged images under PGID-F with different guidance strength $\gamma$. (b) Bit accuracy of forged images under PGID-F with different guidance strength $\gamma$. There is a marginal difference between the performance without guidance and at $\gamma = 0.001$, with the setting $\gamma=0.001$ being slightly better. At higher guidance strengths, the performance became erratic.} 
    \label{fig:combined_forgery_vs_guidance}
\end{figure}

\begin{figure}[t]
    \centering
    \subcaptionbox{\label{fig:dectection_watermarked_vs_guidance}}%
    {\includegraphics[width=0.4\linewidth]{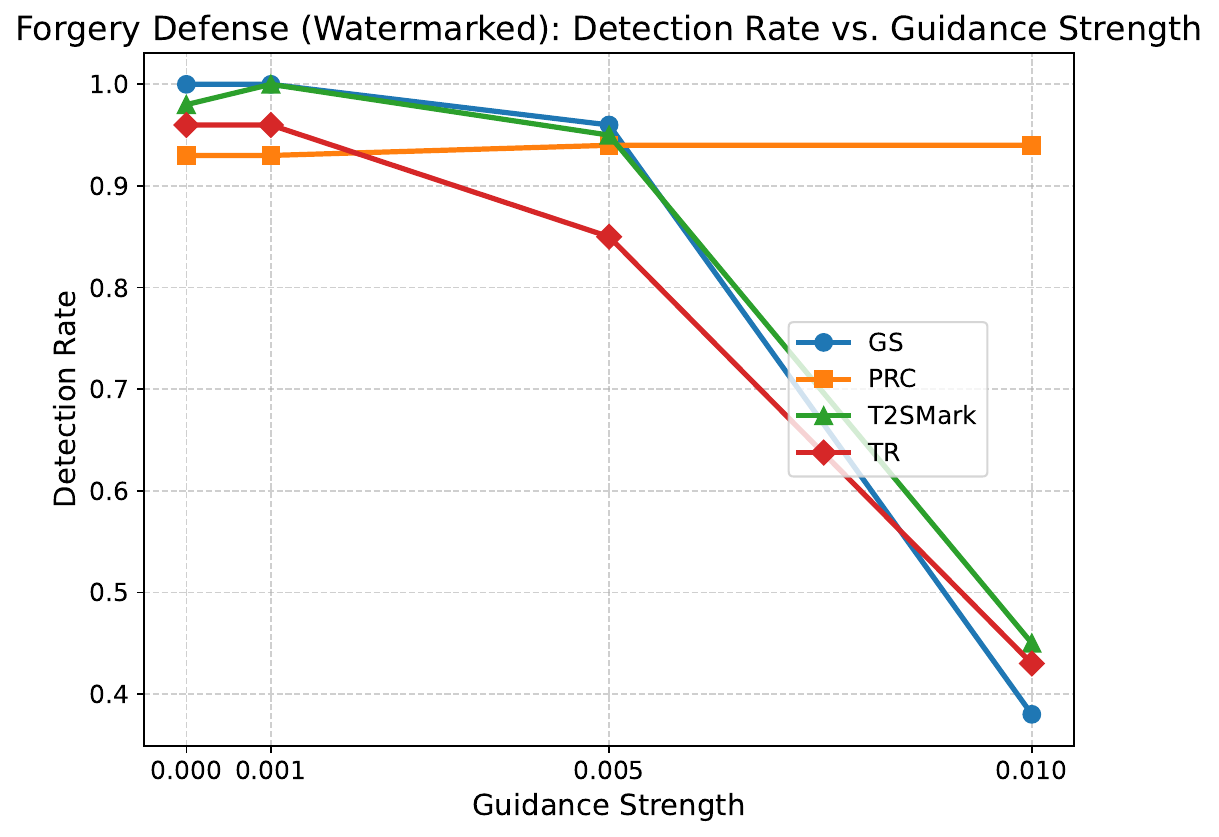}}
    \hspace{0.5cm} 
    \subcaptionbox{\label{fig:bit_watermarked_vs_guidance}}%
    {\includegraphics[width=0.4\linewidth]{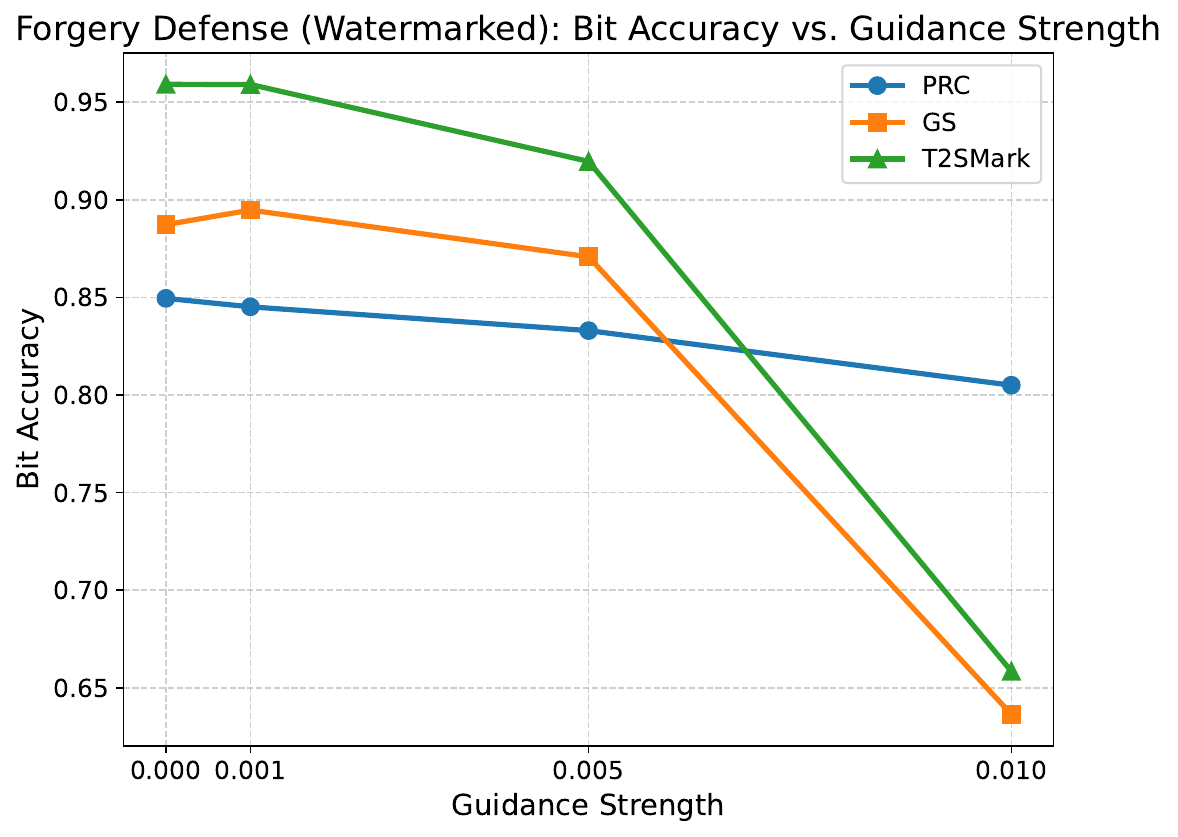}}
    \caption{(a) Detection rate of watermarked images under PGID-F with different guidance strength $\gamma$. (b) Bit accuracy of watermarked images under PGID-F with different guidance strength $\gamma$. Guidance strength $\gamma$ being too high leads to watermarked images losing their watermarks.} 
    \label{fig:combined_watermarked_vs_guidance}
\end{figure}

\textbf{Against forgery attack. }In contrast to the removal case, we observe that the guidance strength $\gamma$ plays a less critical role in defending against forgery. This is because the latent projection of forged images is predominantly driven by the diffusion model's intrinsic manifold projection capabilities. Nevertheless, $\gamma$ provides an additional dimension of control to fine-tune the separation between authentic and forged distributions. Generally, as shown from Tree Ring and PRC in Figure~\ref{fig:combined_forgery_vs_guidance}, higher guidance strength facilitates forged instances to lose watermarks. However, with high guidance strength, some schemes can have erratic behaviors. From Figure~\ref{fig:combined_forgery_vs_guidance}, the detection rate of Gaussian Shading and T2SMark fluctuates as the guidance strength increases. Furthermore, an excessively high guidance strength also risks rejecting authentic watermarks, as shown in Figure~\ref{fig:combined_watermarked_vs_guidance}. 

\subsection{Summary}
Overall, in this section, we first demonstrated the critical effect of skip inversions on the performance of PGID. Then we showed that with a $k$ large enough and suitable $s$ and $\gamma$, one can restore removed watermarks consistently. We also discussed the need for a high $k$ for forgery, and the trade-off based on the robustness of the watermarking scheme used. Next, we discussed the effects of guidance strength, where it is crucial in removal, but not as important in forgery. We want to briefly note that since watermarking schemes behave differently when applied to different model architectures, the working hyperparameter settings of one model might not transfer to another. Finally, we want to mention that in our current evaluation, the thresholds are taken from a desired FPR of the watermarking scheme. However, as shown from the near-perfect AUC in Table~\ref{tab:watermarked_robustness}, the detection scores for forged and authentic watermarks can be well separated with a good choice of the threshold. Therefore, users could choose a threshold for PGID based on their preferred TPR and FPR to classify authentic and forged watermarks.

\section{Computational Analysis}
In this section, we report the runtime comparison between our method and the standard inversion process using DDIM. Our results, as shown in Table~\ref{tab:runtime_benchmark}, indicate that we introduce computational overhead compared to the standard one-stage DDIM inversion. This overhead primarily arises from the progressive latent refinement in the second stage, where a larger $k$ and smaller $s$ result in increased runtime. However, we argue that this increase is a necessary and acceptable trade-off for system security: while standard inversion is faster, it completely fails against the introduced threats. Furthermore, the computational overhead of PGID is highly feasible and manageable, as service providers only need to apply the framework to suspicious images.

\begin{table}[t]
    \input{tables/runtime}
    \vspace{0.4cm}
    \input{tables/DPM_removal}
\end{table}

\section{Against Different Attack Scheduler}

In our primary evaluation, we followed the standard setting of using the DDIM scheduler for the attacker's proxy model. In this section, we consider the case where the attacker utilizes DPM as the scheduler. We report the detection metrics similar to the main experiments. SD 2.1 is employed as both the service provider model and the attacker's proxy model. For cover images in the forgery case, we use the same set of images from MS-COCO as in the main experiments. Our hyperparameter settings remain the same as in our main experiments.

As shown in Table~\ref{tab:dpm_attack_sd21}, Table~\ref{tab:dpm_forgery_attack_sd21}, and Table~\ref{tab:dpm_auc_robustness}, we achieve similar results to when the attacker uses DDIM. PGID-R is still capable of restoring the removed watermarks at an almost perfect detection rate, and PGID-F retains its forgery detection effectiveness by suppressing forged signals to random-guess levels. We conclude that the attacker's scheduler choice does not affect PGID's performance.

\begin{table}[t]
    \input{tables/DPM_forgery}
    \vspace{0.4cm}
    \input{tables/DPM_robustness}
\end{table}

\section{Transferability to Other Attacks}
Even though PGID is designed to be effective against imprint forgery and removal attacks, it could also provide defense against other perturbation-based attacks. In this section, we consider two attacks proposed by Jain et al. (2025)\cite{jain2025forging} and Yang et al. (2024)\cite{yang2024steganalysisdigitalwatermarkingdefense}, which we refer to as the VAE attack and the Averaging attack, respectively. However, it is notable that these two attacks are not as effective and consistent as imprint attacks. From our experiments, we found that the VAE attack only works in the forgery case on three out of the four schemes (Tree Ring, Gaussian Shading, and T2SMark), while the Averaging attack only works on Tree Ring. Previous claims from Yang et al.(2024)\cite{yang2024steganalysisdigitalwatermarkingdefense} about their method working on Gaussian Shading has been disproven in Müller et al. (2025)~\cite{Muller_2025_CVPR}, and we observe similar results. Therefore, we report only the successful attack cases. We will present the results against two attacks: VAE-Forgery on three schemes and Average-Removal on Tree Ring.

\textbf{VAE-Forgery. }Jain et al.(2025)\cite{jain2025forging} proposed forging the watermarks from a single watermarked image $x^{(w)}$ to a cover image $x^{(c)}$ by optimizing a perturbation $\delta$ in the pixel space with the following loss function:
\begin{equation} 
    \mathcal{L}_{\text{forgery}} = \|\mathcal{E} ({x}^{(c)} + \delta) - \mathcal{E}({x}^{(w)}) \|_2 + \lambda \| \delta \|_2
\end{equation}
where $\mathcal{E}$ is a proxy encoder used by the attacker. The attack aims to push the latent representation of the cover image closer to a watermarked image. We follow the setting from the paper where $\lambda = 5 \times 10^4$ and perform $100$ optimization steps. We follow the implementation in the paper where the proxy VAE from SD 1.4 is used.\footnote{https://huggingface.co/CompVis/stable-diffusion-v1-4}

\textbf{Average-Removal. }Yang et al.(2024)\cite{yang2024steganalysisdigitalwatermarkingdefense} found that for many watermarking schemes, the watermark signal can be extracted by averaging the residual across many watermarked and unwatermarked images. In their method, the watermark signal can be extracted as:
\begin{equation}
    \delta = \frac{1}{N} \left( \sum_{i=1}^{N} x_{w,i} - \sum_{i=1}^{N} x_{c,i} \right),
\end{equation}
where $N$ is the number of averaged images, $\{x_{c,i}\}_{i \in [N]}$ are clean unwatermarked images, and $\{x_{w,i}\}_{i \in [N]}$ are watermarked images.
To perform removal, they subtract the extracted watermark pattern from the watermarked images to create the attacked image $x^{atk}$:
\begin{equation}
    x^{atk} = x^{(w)} - \delta
\end{equation}
For our evaluation, we averaged over 5,000 watermarked images to extract the watermark pattern. We consider the stronger grey-box scenario of the method, where the extracted pattern is calculated from paired watermarked and unwatermarked images.

\textbf{Setup and Results. }We utilize SD 2.1 as the service provider model and use the same set of hyperparameters as in the main experiments. Metrics are evaluated on $100$ images of each type, similar to the main experiments.
\begin{itemize}
    \item Against VAE-Forgery: As presented in Table~\ref{tab:vae_forgery_attack} and Table~\ref{tab:vae_auc_robustness}, PGID-F successfully reduces the detection rate of forged instances across all watermarking methods. The AUC and bit accuracy also drop, indicating effective defense against this forgery attack.
    \item Against Average-Removal: As shown in Table~\ref{tab:averaging_attack}, PGID-R successfully restores the watermarks for Tree Ring, indicating our method's potential for defending other perturbation-based removal attacks. 
\end{itemize}

\input{tables/vae_forgery}

\begin{table}[t]
    \input{tables/vae_robustness}
    \vspace{0.4cm}
    \input{tables/averaging}
\end{table}

\section{Evaluation on Larger Test Sets}

To further assess the robustness of PGID, we extend our evaluation
to 400 images per category for Tree-Ring and Gaussian Shading,
quadrupling the original test set size.
We follow the same experimental settings as in our main experiments,
except for the increased test set size.
Tables \ref{tab:removal_attack_400} and
\ref{tab:forgery_attack_400} report the results against imprint
removal and forgery attacks, respectively.
For the forgery evaluation, AUC is calculated between watermarked
and forged images.
The results demonstrate that PGID maintains strong performance
on the larger test set, confirming its effectiveness as the
evaluation set size increases.

\input{tables/rebuttal_scale}

\section{Limitations}
While PGID significantly enhances the robustness of semantic watermark detection, we acknowledge a few limitations in our current framework. First, the hyperparameter configurations utilized in our evaluations were determined empirically through trial and error, meaning they may not represent the strictly optimal settings. Second, optimal hyperparameter settings might vary depending on the watermarking scheme and model architecture used by the service provider. As mentioned in Section~\ref{appendix:hyperparam_settings}, watermarking schemes with high sensitivity and low robustness, such as PRC, might require more careful and extensive hyperparameter tuning. Finally, because PGID is fundamentally designed to mitigate perturbations, it currently cannot protect against perturbation-free attacks such as Rinse~\cite{an2024benchmarking} or the recent next-frame prediction attack~\cite{qiu2025the}.

%% file: tables/runtime.tex
\centering
\footnotesize
\caption{Runtime benchmark in milliseconds (ms). We compare standard DDIM inversion to PGID across three stages with the settings in our main results. All of DDIM, Stage I, and Stage III use $50$ inference steps.}
\vspace{1ex}
\label{tab:runtime_benchmark}
\begin{tabular}{@{}lcccc@{}}
\toprule
\textbf{Pipeline / Stage} & \textbf{Stage I} & \textbf{Stage II} & \textbf{Stage III} & \textbf{Total (ms)} \\
\midrule
DDIM & -- & -- & -- & 2039 \\
PGID-R ($k=10$, $s=1$) & 2039 & 4086 & 2040 & 8165 \\
PGID-F ($k=15$, $s=3$) & 2040 & 8092 & 2041 & 12172 \\
PGID-F ($k=7$, $s=3$) & 2039 & 1553 & 2039 & 5631 \\
\bottomrule
\end{tabular}

%% file: tables/DPM_removal.tex
\centering
\caption{Results against imprint removal attack when the attacker utilizes the DPM scheduler. Results are reported for the attack at 50/100/150 optimization steps. Higher AUC and Det. indicate higher robustness to the attack. The best result for each scheme is in \textbf{bold}.}
\vspace{1ex}
\label{tab:dpm_attack_sd21}
\setlength{\tabcolsep}{2mm}
\resizebox{0.75\textwidth}{!}{
\begin{tabular}{@{}l ccc @{}}
\toprule
\multirow{2}{*}{Method} & \multicolumn{3}{c}{Attacker Scheduler: DPM (SD 2.1)} \\
\cmidrule(l){2-4}
& Det.$\uparrow$ & Bit Acc.$\uparrow$ & AUC$\uparrow$ \\
\midrule
TR        & 0.25/0.13/0.10 & --- & 0.4033/0.1877/0.1384 \\
\rowcolor{gray!15}
TR+P      & \textbf{1.00}/\textbf{1.00}/\textbf{1.00} & --- & \textbf{0.9998}/\textbf{1.0000}/\textbf{1.0000} \\
GS        & 0.00/0.00/0.00 & 0.2586/0.0326/0.0062 & 0.0734/0.0000/0.0000 \\
\rowcolor{gray!15}
GS+P      & \textbf{1.00}/\textbf{1.00}/\textbf{1.00} & \textbf{0.9979}/\textbf{0.9990}/\textbf{0.9990} & \textbf{1.0000}/\textbf{1.0000}/\textbf{1.0000} \\
PRC       & 0.00/0.00/0.00 & 0.4976/0.4994/0.4999 & --- \\
\rowcolor{gray!15}
PRC+P     & \textbf{0.97}/\textbf{1.00}/\textbf{0.99} & \textbf{0.8748}/\textbf{1.0000}/\textbf{0.9753} & --- \\
T2SMark   & 0.49/0.95/\textbf{1.00} & 0.5111/0.4945/0.4975 & 0.8481/0.9993/\textbf{1.0000} \\
\rowcolor{gray!15}
T2SMark+P & \textbf{1.00}/\textbf{1.00}/\textbf{1.00} & \textbf{0.9998}/\textbf{1.0000}/\textbf{1.0000} & \textbf{1.0000}/\textbf{1.0000}/\textbf{1.0000} \\
\bottomrule
\end{tabular}}

%% file: tables/DPM_forgery.tex
\centering
\caption{Results against imprint forgery attack on MS-COCO dataset when the attacker utilizes the DPM scheduler. Results are reported for the attack at 50/100/150 optimization steps. Lower AUC and Det. indicate higher robustness to the attack. The best result for each scheme is in \textbf{bold}.}
\vspace{1ex}
\label{tab:dpm_forgery_attack_sd21}
\setlength{\tabcolsep}{2mm}
\resizebox{0.75\textwidth}{!}{
\begin{tabular}{@{}l ccc @{}}
\toprule
\multirow{2}{*}{Method} & \multicolumn{3}{c}{Attacker Scheduler: DPM (SD 2.1)} \\
\cmidrule(l){2-4}
& Det.$\downarrow$ & Bit Acc.$\downarrow$ & AUC$\downarrow$ \\
\midrule
TR        & 1.00/1.00/1.00 & --- & 1.0000/1.0000/1.0000 \\
\rowcolor{gray!15}
TR+P      & \textbf{0.01}/\textbf{0.07}/\textbf{0.10} & --- & \textbf{0.5485}/\textbf{0.6111}/\textbf{0.6651} \\
GS        & 1.00/1.00/1.00 & 0.9992/0.9995/0.9996 & 1.0000/1.0000/1.0000 \\
\rowcolor{gray!15}
GS+P      & \textbf{0.00}/\textbf{0.00}/\textbf{0.02} & \textbf{0.5396}/\textbf{0.5781}/\textbf{0.6093} & \textbf{0.7842}/\textbf{0.9203}/\textbf{0.9599} \\
PRC       & 0.96/1.00/1.00 & 0.8964/1.0000/1.0000 & --- \\
\rowcolor{gray!15}
PRC+P     & \textbf{0.00}/\textbf{0.00}/\textbf{0.03} & \textbf{0.4964}/\textbf{0.5041}/\textbf{0.4951} & --- \\
T2SMark   & 1.00/1.00/1.00 & 1.0000/1.0000/1.0000 & 1.0000/1.0000/1.0000 \\
\rowcolor{gray!15}
T2SMark+P & \textbf{0.00}/\textbf{0.00}/\textbf{0.01} & \textbf{0.4982}/\textbf{0.5072}/\textbf{0.5141} & \textbf{0.5294}/\textbf{0.6645}/\textbf{0.7657} \\
\bottomrule
\end{tabular}}

%% file: tables/DPM_robustness.tex
\centering
\caption{Separability of authentic watermarks against forgery attack using the DPM scheduler. AUC is calculated between watermarked images and forged images for the attacks at 50, 100, and 150 optimization steps (50/100/150).}
\vspace{1ex}
\label{tab:dpm_auc_robustness}
\begin{tabular}{@{}l c@{}}
\toprule
\multirow{2}{*}{Method} & Attacker Scheduler: DPM (SD 2.1) \\
\cmidrule(l){2-2}
& AUC$\uparrow$\\
\midrule 
TR+P      & 0.9924/0.9887/0.9855 \\
GS+P      & 1.0000/1.0000/1.0000 \\
PRC+P     & --- \\
T2SMark+P & 1.0000/1.0000/0.9998 \\
\bottomrule
\end{tabular}

%% file: tables/vae_forgery.tex
\begin{table}[t]
\centering
\caption{Results against VAE-Forgery attack on MS-COCO dataset. The attacker optimizes the pixel-space perturbation for 100 steps using the SD 1.4 proxy VAE. Lower AUC and Det. indicate higher robustness to the attack. The best result for each scheme is in \textbf{bold}.}
\vspace{1ex}
\label{tab:vae_forgery_attack}
\setlength{\tabcolsep}{2mm}
\begin{tabular}{@{}l ccc @{}}
\toprule
\multirow{2}{*}{Method} & \multicolumn{3}{c}{Attacker Proxy VAE: SD 1.4} \\
\cmidrule(l){2-4}
& Det.$\downarrow$ & Bit Acc.$\downarrow$ & AUC$\downarrow$ \\
\midrule
TR        & 0.77 & ---    & 0.9565 \\
\rowcolor{gray!15}
TR+P      & \textbf{0.27} & ---    & \textbf{0.7326} \\
GS        & 0.99 & 0.9359 & 1.0000 \\
\rowcolor{gray!15}
GS+P      & \textbf{0.02} & \textbf{0.6138} & \textbf{0.9866} \\
T2SMark   & 1.00 & 0.9814 & 1.0000 \\
\rowcolor{gray!15}
T2SMark+P & \textbf{0.02} & \textbf{0.5156} & \textbf{0.5744} \\
\bottomrule
\end{tabular}
\end{table}

%% file: tables/vae_robustness.tex
\centering
\caption{Separability of authentic watermarks against VAE-forgery attack. AUC is calculated between watermarked images and forged images.}
\vspace{1ex}
\label{tab:vae_auc_robustness}
\begin{tabular}{@{}l c@{}}
\toprule
\multirow{2}{*}{Method} & Attacker Proxy VAE: SD 1.4 \\
\cmidrule(l){2-2}
& AUC$\uparrow$\\
\midrule 
TR+P      & 0.9381 \\
GS+P      & 0.9996 \\
T2SMark+P & 0.9995 \\
\bottomrule
\end{tabular}

%% file: tables/averaging.tex
\centering
\small
\caption{Results against the Average-Removal attack on the Tree Ring. PGID effectively recovers the removed watermarks, as indicated by the restored detection rate and high AUC.}
\vspace{1ex}
\label{tab:averaging_attack}
\setlength{\tabcolsep}{3mm}
\begin{tabular}{@{}l cc@{}}
\toprule
Method & Det.$\uparrow$ & AUC$\uparrow$ \\
\midrule
TR        & 0.10          & 0.1460 \\
\rowcolor{gray!15}
TR+P      & \textbf{0.97} & \textbf{0.9885} \\
\bottomrule
\end{tabular}

%% file: tables/rebuttal_scale.tex
\begin{table}[t]
\centering

\begin{minipage}[t]{0.49\textwidth}
\centering
\caption{Results against imprint removal attacks on 400 images per category. Results are reported for the attack at 50/100/150 optimization steps (Attacker Model: SD 2.1).}
\label{tab:removal_attack_400}
\vspace{1ex}
\setlength{\tabcolsep}{1mm}
\resizebox{\linewidth}{!}{
\begin{tabular}{@{}l ccc@{}}
\toprule
Method & Det.$\uparrow$ & Bit Acc.$\uparrow$ & AUC$\uparrow$ \\
\midrule
TR+P & 0.9975/0.9975/1.0000 & --- & 0.9999/0.9998/0.9999 \\
GS+P & 1.0000/1.0000/1.0000 & 0.9970/0.9980/0.9977 & 1.0000/1.0000/1.0000 \\
\bottomrule
\end{tabular}}
\end{minipage}
\hfill
\begin{minipage}[t]{0.49\textwidth}
\centering
\caption{Results against imprint forgery attacks on 400 images per category. Results are reported for the attack at 50/100/150 optimization steps (Attacker Model: SD 2.1).}
\label{tab:forgery_attack_400}
\vspace{1ex}
\setlength{\tabcolsep}{1mm}
\resizebox{\linewidth}{!}{
\begin{tabular}{@{}l ccc@{}}
\toprule
Method & Det.$\downarrow$ & Bit Acc.$\downarrow$ & AUC$\uparrow$ \\
\midrule
TR+P & 0.0375/0.0650/0.1100 & --- & 0.9912/0.9873/0.9826 \\
GS+P & 0.0025/0.0025/0.0325 & 0.5414/0.5815/0.6133 & 0.9998/0.9996/0.9990 \\
\bottomrule
\end{tabular}}
\end{minipage}

\end{table}